\documentclass{article}

 \usepackage[preprint]{neurips_2026}
 
\usepackage[utf8]{inputenc} 
\usepackage[T1]{fontenc}    
\usepackage{hyperref}       
\usepackage{url}            
\usepackage{booktabs}       
\usepackage{amsfonts}       
\usepackage{nicefrac}       
\usepackage{microtype}      
\usepackage{xcolor}         
\usepackage{amsmath}
\usepackage{amssymb}
\usepackage{mathtools}
\usepackage{amsthm}
\usepackage{microtype}
\usepackage{graphicx}
\usepackage{subcaption}
\usepackage{booktabs} 
\usepackage{multirow}
\usepackage{algorithm}
\usepackage{algorithmic}
\usepackage{wrapfig}
\usepackage{amsthm}

\usepackage{placeins}

\title{Learning Defensive Policies against Diverse Inference Attacks for Smart Meter Privacy}

\author{%
  Ruichang Zhang\textsuperscript{1}
  \qquad
  Mustafa A. Mustafa\textsuperscript{1,2}
  \\[2mm]
  \textsuperscript{1}  Department of Computer Science, The University of Manchester, Manchester, UK
  \\
  \textsuperscript{2}COSIC, KU Leuven, Leuven, Belgium
}

\begin{document}

\maketitle

\begin{abstract}

Smart meter (SM) data provides fine-grained visibility into household energy consumption, but also exposes users to privacy risks. 
Inference attacks, known as non-intrusive load monitoring (NILM), can perform appliance-level inference from aggregate signals and recover sensitive behavioral patterns.
In practice, attacker models are unknown and heterogeneous, making robust defense challenging. 
We formulate SM privacy protection as a black-box inference defense problem, aiming to reduce the recoverability of appliance-level information while generalizing across diverse and unseen attackers. 
We propose a proxy-guided hierarchical reinforcement learning framework that learns battery-based load-shaping policies to inject realistic but misleading appliance-level signatures into the aggregate signal, thereby disrupting the structured patterns exploited by NILM. A self-supervised aggregate-structure privacy probe provides a reconstruction-error-based surrogate reward for disrupting recoverable load structure, while a signature library makes the perturbations appliance-relevant and physically realizable through battery control. 
We provide theoretical rationale showing that proxy-guided optimization improves inference robustness under attacker diversity.
Experiments on real-world datasets UK-DALE and REDD demonstrate strong cross-model and cross-appliance generalization.
Across six unseen NILM attackers, covering four appliances on UK-DALE and five on REDD, our proposed defense increases average appliance-level RMSE by 107\% and 166\%, respectively, while reducing F1 score by 79\% and 80\%.

\end{abstract}

\section{Introduction}
Smart meters (SMs) record high-frequency, real-time electricity consumption time series, where aggregate load measurements encode structured appliance-level usage patterns that can be exploited by non-intrusive load monitoring (NILM) models~\citep{hart1992nonintrusive}.
Prior work~\cite{kolter2012approximate,kelly2015neural,chen2018convolutional,zhang2018sequence,yue2020bert4nilm,sykiotis2022electricity} has shown that diverse NILM methods can recover fine-grained appliance activities from aggregate signals without direct access to individual devices.
Such appliance-level information reveals latent household behaviors and therefore raises serious privacy concerns.


Existing SM privacy defenses largely optimize proxy objectives that are not explicitly aligned with the NILM disaggregation task.
For example, prior battery-based or load-shaping methods often encourage curve smoothness~\cite{erdemir2020privacy,zhang2024privacy}, reduce aggregate variance~\cite{li2023research,zhang2024proactive}, optimize other load-level proxies~\cite{farokhi2019fisher,farokhi2020fundamental}, or rely on information-theoretic criteria~\cite{shateri2021privacy,shateri2023privacy}.
While these objectives can reshape aggregate trajectories, they are not explicitly aligned with the inference structure exploited by NILM attackers.
As a result, their influence on appliance-level inference is only indirectly controlled, creating a privacy–task mismatch between the defender's training signal and the attacker's actual inference goal. 
The key challenge is therefore to learn defender policies that disrupt NILM-relevant load structures while remaining robust to diverse and unseen attackers.

To address this challenge, we propose a proxy-guided appliance-signature mimicry framework that combines aggregate-structure disruption with a NILM-relevant signature-level intervention mechanism.
Specifically, we use the reconstruction error of an aggregate-level sequence probe network as the training signal, encouraging the defender to disrupt recognizable load structures that NILM models rely on for appliance-level disaggregation.
The probe is trained in a self-supervised aggregate-to-aggregate manner and does not require appliance-level labels or access to the deployed NILM attacker.
Meanwhile, instead of injecting arbitrary perturbations, the defender inserts realistic appliance signatures drawn from a curated signature library through feasible battery control, thereby disrupting the appliance-level evidence exploited by NILM models.
We formulate SM privacy protection as a black-box inference defense problem under attacker uncertainty, where the defender does not know the exact NILM model used during deployment.
To solve this problem, we learn a hierarchical reinforcement learning (HRL) policy that decides when and which appliance signature to mimic, while a low-level battery executor realizes the selected signature under physical constraints.
The resulting policy reshapes the reported aggregate load with realistic appliance-like patterns to reduce the recoverability of true appliance-level consumption.
We evaluate the proposed framework on two benchmark SM datasets (UK-DALE and REDD).
Experiments show that the proposed framework consistently degrades NILM reconstruction performance across a diverse set of unseen NILM attackers and remains effective across different datasets, households, and seasonal conditions.

\textbf{Contributions.} Our contributions can be summarized as follows:
(1) We propose a privacy defense framework that injects realistic appliance signatures into aggregate load signals via a curated signature library to mask appliance-level consumption patterns.
(2) We develop a proxy-guided hierarchical reinforcement learning (HRL) framework that learns a battery control policy using an aggregate-level privacy probe as the training signal.
(3) We formulate SM privacy as a black-box inference defense problem under attacker uncertainty. We theoretically justify proxy-guided signature manipulation through the unified task formulation of NILM inference and the adversarial transferability of surrogate-based attacks.
(4) We empirically demonstrate consistent degradation of attacker inference performance across multiple NILM models and datasets, including unseen attackers. 
The results show consistent effectiveness across different datasets and strong generalization across households and seasonal conditions.


\section{Related work}
\label{sec:related work}

\paragraph{NILM as inference threat.}
NILM methods recover appliance-level consumption from aggregate SM signals, ranging from combinatorial optimization~\cite{hart1992nonintrusive} and factorial HMMs~\cite{kolter2012approximate} to convolutional, recurrent, and transformer-based sequence models~\cite{kelly2015neural,zhang2018sequence,yue2020bert4nilm,sykiotis2022electricity,petralia2025nilmformer}.
Despite this architectural diversity, recent surveys~\cite{rafiq2024review,kaselimi2022towards} note that modern NILM models share a common task formulation: estimating appliance-level power or on/off state from a window of aggregate readings, evaluated under a shared family of reconstruction losses.
This task-level commonality, combined with architectural diversity, motivates the black-box defense setting in this work, where the defender does not know the deployed NILM architecture but can exploit the shared structure of the inference task.

\paragraph{SM privacy defenses.}
Existing SM privacy approaches fall into data-centric and physical-layer categories.
Data-centric mechanisms such as aggregation~\cite{gomez2014smart} and differential privacy~\cite{zhang2025user} perturb or post-process reported readings, but implicitly assume that adversaries cannot tap the physical power line, which limits their applicability under realistic deployment.
Physical-layer approaches reshape household consumption with a rechargeable battery before measurement~\cite{erdemir2019privacy,shateri2020privacy}, providing local and real-time protection against attackers that exploit the structure of real load trajectories.
Early battery-based methods rely on rule-based or heuristic strategies~\cite{erdemir2019privacy,farokhi2019fisher,farokhi2020fundamental}, while more recent work introduces RL to automate battery scheduling~\cite{shateri2020privacy,shateri2021privacy,erdemir2020privacy,li2023research,shateri2023privacy,zhang2024proactive,zhang2024privacy}.
These RL-based methods typically optimize either load-level proxies such as smoothness~\cite{erdemir2020privacy,zhang2024privacy} or variance reduction~\cite{li2023research,zhang2024proactive}, or information-theoretic measures such as mutual information and information leakage~\cite{shateri2021privacy,shateri2023privacy}.
Load-level proxies do not explicitly model how attackers exploit appliance-level temporal patterns, leading to a privacy--inference gap when attackers rely on different representations.
Information-theoretic objectives are conceptually principled but cannot be directly attributed to individual control actions, and existing schemes therefore rely on auxiliary estimators whose transferability to unseen attackers remains unclear.

\paragraph{Adversarial robustness and transferability.}
Our formulation also connects to transfer-based black-box adversarial attacks, where perturbations crafted on a surrogate model transfer to unseen targets~\citep{papernot2017practical,liu2016delving,gu2023survey}.
Work~\citep{demontis2019adversarial} formally analyze attack transferability and identify three governing factors: the target model's intrinsic adversarial vulnerability, the surrogate's complexity, and gradient alignment between surrogate and target.
While most analyses focus on classification, recent work shows that surrogate-based attacks also transfer in time-series regression~\citep{mode2020adversarial}, and that selective, confidence-and-error-aware perturbations outperform unstructured ones under fixed budgets~\citep{tokgoz2026intarg}.
This motivates using a surrogate privacy probe to measure load-structure recoverability and guide the learning of transferable battery-based perturbation policies.
Our work instantiates these principles in a physically constrained, RL-controlled setting, where perturbations are realized through battery actions and structured as appliance-signature mimicry.

\begin{figure}[tbp]
  \centering
  \includegraphics[scale=0.12]{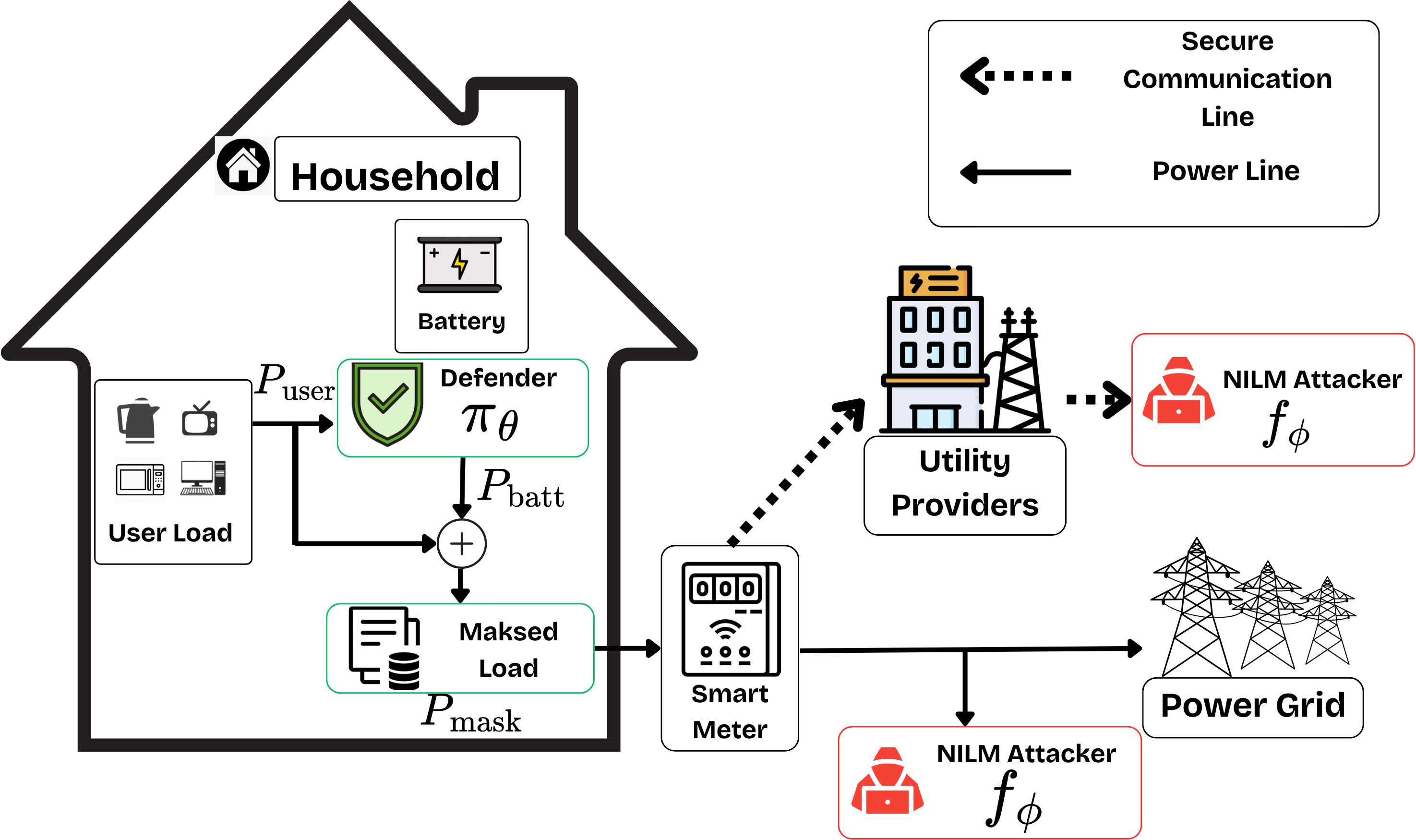}
  \caption{Defender $\pi_\theta$ injects battery-driven mimic signatures $P_{\text{batt}}$ into user load signal $P_{\text{user}}$, aiming to deceive an NILM attacker $f_\phi$ who attempts to infer appliance patterns from aggregated SM data.}
  \label{fig:system}
\end{figure}

\section{Problem Formulation}
\label{sec:system_model}
We consider a black-box SM privacy setting in which the defender does not know the deployed NILM attacker (see Fig.~\ref{fig:system}).
At time $t$, let $P_{\mathrm{user}}(t)$ denote the original household aggregate load before defense, and let $P_{\mathrm{batt}}(t)$ denote the battery power controlled by the defender.
We follow the convention that $P_{\mathrm{batt}}(t)>0$ denotes charging and $P_{\mathrm{batt}}(t)<0$ denotes discharging.
The smart meter reports the masked aggregate load $P_{\mathrm{mask}}^{\pi_\theta}(t)=P_{\mathrm{user}}(t)+P_{\mathrm{batt}}(t)$ where the battery action is generated by a defender policy $\pi_\theta$ and must satisfy physical constraints, including capacity limits, charging and discharging power limits, state-of-charge feasibility, and throughput constraints.
For a window length $\tau$, we denote the clean aggregate window by $\mathbf{x}^{0}_{t}=[P_{\mathrm{user}}(t-\tau+1),\ldots,P_{\mathrm{user}}(t)]$ and the corresponding masked aggregate window induced by policy $\pi_\theta$ by $\mathbf{x}^{\pi_\theta}_{t}=[P_{\mathrm{mask}}^{\pi_\theta}(t-\tau+1),\ldots,P_{\mathrm{mask}}^{\pi_\theta}(t)]$. 
The appliance-level target within the same window is denoted by $\mathbf{y}_{t}$.

An NILM attacker $f_\phi$ maps the reported aggregate to appliance-level predictions:
$\hat{\mathbf{y}}_{t}=f_\phi(\mathbf{x}^{\pi_\theta}_{t}). $
For a given attacker $f_\phi$ and defender policy $\pi_\theta$, we define the attacker's inference risk as
\begin{equation}
\label{eq:attacker_inference_risk}
R(f_\phi,\pi_\theta)
    =
    \mathbb{E}_{t}
    \left[
        \mathcal{L}
        \left(
            f_\phi(\mathbf{x}_{t}^{\pi_\theta}),
            \mathbf{y}_{t}
        \right)
    \right],
\end{equation}
where $\mathcal{L}(\cdot)$ denotes appliance-level reconstruction loss, such as RMSE, MAE, or SAE.
For activation-state inference, we additionally report detection degradation using F1 reduction.
A larger $R(f_\phi,\pi_\theta)$ indicates that appliance-level consumption is harder to recover from the reported aggregate signal, and therefore corresponds to stronger privacy protection.
Since the deployed NILM attacker is unknown, we assume that the attacker belongs to a heterogeneous family $f_\phi\in\mathcal{F}$, covering optimization-based, probabilistic, and neural sequence prediction models.
The ideal robust privacy objective is to maximize the expected NILM inference risk over this attacker family:
\begin{equation}
\label{eq:robust_objective}
\max_{\theta}
    \quad
    \mathbb{E}_{f_\phi \sim \mu}
    \left[
        R(f_\phi,\pi_\theta)
    \right],
\end{equation}
where $\mu$ denotes an unknown distribution over possible NILM attackers. 
Direct optimization of Eq.~\eqref{eq:robust_objective} is infeasible because the deployed attacker and the distribution $\mu$ are not accessible during training.
We therefore introduce a self-supervised aggregate-structure privacy probe in Section~\ref{sec:theoretical_rationale} to provide a proxy learning signal.

\section{Theoretical Rationale}
\label{sec:theoretical_rationale}
The black-box defense problem in Section~\ref{sec:system_model} can be viewed as a physically constrained adversarial attack against NILM inference models, where the perturbation set $\Delta_{\mathrm{batt}}$ is determined by feasible battery actions rather than norm-bounded digital noise.
We denote the ideal expected inference risk over the attacker family by
$\mathcal{R}_{\mathrm{avg}}(\pi_\theta) =
\mathbb{E}_{f_\phi \sim \mu}[R(f_\phi,\pi_\theta)]$.
This quantity is the robust privacy objective in Eq.~\eqref{eq:robust_objective}.
We therefore develop a computable proxy that serves as a tractable surrogate for $\mathcal{R}_{\mathrm{avg}}(\pi_\theta)$, resting on two design choices: an aggregate-level privacy probe as the surrogate training signal, and signature mimic as the perturbation mechanism.

\paragraph{Proxy-guided adversarial optimization.}
We introduce an aggregate-level privacy probe $f_\eta$ trained on aggregate load data via identity self-supervision: the probe takes an aggregate window $\mathbf{x}^{0}$ as input and is trained to reconstruct $\mathbf{x}^{0}$ as output.
The probe is kept fixed during defender training.
Formally, we define the proxy inference risk as
\begin{equation}
\label{eq:proxy_inference_risk}
    R_p(\pi_\theta)
    =
    \mathbb{E}_{t}
    \left[
        \mathcal{L}
        \left(
            f_\eta(\mathbf{x}^{\pi_\theta}_{t}),
            \mathbf{x}^{0}_{t}
        \right)
    \right].
\end{equation}
The defender is then trained to increase this proxy inference risk:
\begin{equation}
\label{eq:proxy_guided_objective}
    \max_{\theta}
    \ R_p(\pi_\theta).
\end{equation}
Because $f_\eta$ is trained to reconstruct clean aggregate windows, 
a larger $R_p(\pi_\theta)$ indicates that the masked aggregate window deviates from the clean aggregate patterns learned by the probe.
Thus, $R_p$ measures the degree to which the defender disrupts aggregate-level temporal regularities, rather than directly estimating appliance-level NILM error.

We use $R_p(\pi_\theta)$ as a computable surrogate for $\mathcal{R}_{\mathrm{avg}}(\pi_\theta)$ during training.
The surrogate replaces (i) the appliance-level target $\mathbf{y}_t$ with the aggregate $\mathbf{x}^0_t$, 
which is available without any appliance-level supervision, 
and (ii) the unknown attacker distribution $\mu$ with the fixed probe $f_\eta$. 
The rationale for this substitution is that NILM attackers share a common dependence on aggregate temporal patterns, which the probe is designed to capture.

\paragraph{Task commonality across NILM attackers.}
Although NILM attackers differ substantially in formulation—from optimization-based disaggregation~\citep{hart1992nonintrusive} and probabilistic state-space models~\citep{kolter2012approximate} to neural sequence  predictors~\citep{kelly2015neural}—they share a common premise: appliance-level outputs are inferred from aggregate signals under the assumption that the observed aggregate is consistent with the aggregate–appliance co-occurrence statistics seen during training or specified as a generative model.
Optimization-based methods search for sparse signature combinations that explain the aggregate; probabilistic models treat the aggregate as an emission from a hidden appliance state with distribution learned from data; neural predictors learn a direct mapping from aggregate trajectories to appliance outputs. In all cases, inference degrades when the reported aggregate is pushed 
into a region where these training-time statistics no longer hold.
Our probe $f_\eta$, trained to reconstruct clean aggregate windows from the same data distribution,  serves as a tractable summary of these statistics. A large $R_p(\pi_\theta)$ therefore certifies that $\mathbf{x}^{\pi_\theta}_t$ has been driven into such a region, even though the probe itself does not perform appliance-level inference.
Pushing the masked aggregate away from these shared regularities is therefore expected to increase the inference risk of many unseen attackers in $\mathcal{F}$, providing a tractable route for increasing $\mathcal{R}_{\mathrm{avg}}(\pi_\theta)$.

This argument aligns with the surrogate-model principle of black-box adversarial attacks~\citep{papernot2017practical,liu2016delving,gu2023survey}: perturbations optimized against a fixed reference model transfer to other models that share its training data and inductive biases.
Work\citep{demontis2019adversarial} formally show that, under linearization of the attack loss, transferability is governed by the target's intrinsic vulnerability, the surrogate's complexity, and the gradient alignment between surrogate and target.
Work~\citep{mode2020adversarial} provide empirical evidence that the principle extends to time-series regression: adversarial perturbations crafted on a CNN regressor transfer with substantial reconstruction-error increase across LSTM and GRU regressors on power consumption data.
The structural condition under which this transfer provably increases $\mathcal{R}_{\mathrm{avg}}$ is given in Appendix~\ref{app:theory_analysis}.

\paragraph{Signature mimic as a task-aligned perturbation.}
Since NILM attackers recover appliance-level consumption from aggregate load, an effective perturbation interferes with the appliance-level evidence used for disaggregation rather than merely adding aggregate noise.
We construct perturbations from a library $\mathcal{S}$ of real appliance signatures:
\begin{equation}
\label{eq:signature_perturbation_main}
    \mathbf{x}^{\pi_\theta}_{t}
    =
    \mathbf{x}^{0}_{t}
    +
    \delta_{\mathrm{sig}},
    \qquad
    \delta_{\mathrm{sig}}
    \in
    \Delta_{\mathrm{sig}},
\end{equation}
where $\delta_{\mathrm{sig}}$ is induced by feasible battery actions that replay a selected signature trajectory.

The perturbation is therefore physically realizable while retaining appliance-like temporal structure.
By restricting the policy to signature-induced perturbations, the framework converts aggregate-structure disruption into plausible but misleading appliance-level evidence, thereby creating ambiguous disaggregation explanations for NILM attackers.
This view is consistent with a broader trend in time-series adversarial attacks: informed, selective perturbations that exploit task structure outperform unstructured ones under fixed budgets~\citep{tokgoz2026intarg}.
In SM data privacy tasks, signature-based interventions provide a structured perturbation class better matched to the NILM inference task.
The next section instantiates these principles in a hierarchical RL framework operationalizing proxy-guided signature mimic under realistic battery constraints.
\section{Proxy-Guided HRL Policy Training}
\label{sec:methodology}
We instantiate the proxy-guided signature mimicry principle in Section~\ref{sec:theoretical_rationale} as a hierarchical RL framework: the fixed privacy probe provides the reward signal, while the signature library defines the high-level perturbation action space.
The defender policy \(\pi_\theta\) learns when to apply signature mimicry and which appliance-like trajectory to replay, while all battery actions are executed under hard physical feasibility constraints.

\paragraph{Proxy-based privacy reward.}
The training objective in Eq.~\eqref{eq:proxy_guided_objective} maximizes the expected proxy 
inference risk $R_p(\pi_\theta)$. We optimize this objective with PPO by deriving a per-step 
privacy reward from its instantaneous form. Since the probe $f_\eta$ has nonzero residual 
reconstruction error even on clean aggregate windows in practice, we measure the defender's 
contribution as the \emph{increase} in probe error caused by signature insertion, relative to 
the probe's residual on the unmodified window $\mathbf{x}^{0}_{t}$:

\begin{equation}
\label{eq:privacy_reward}
    r_t^{\mathrm{priv}}
    =
    \mathcal{L}
    \left(
        f_\eta(\mathbf{x}^{\pi_\theta}_{t}),
        \mathbf{x}^{0}_{t}
    \right)
    -
    \mathcal{L}
    \left(
        f_\eta(\mathbf{x}^{0}_{t}),
        \mathbf{x}^{0}_{t}
    \right).
\end{equation}
Thus, the policy receives positive reward when the selected signature insertion makes the reported aggregate harder to reconstruct as the original clean aggregate structure under the fixed privacy probe. 
Battery operation also incurs both monetary energy cost $C_{\text{energy}}(t)$ and battery degradation cost $C_{\text{deg}}(t)$. Hence we use the combined reward 
\begin{equation} 
\label{eq:total_reward} 
r_t = \lambda r_t^{\mathrm{priv}} - (1-\lambda) \left( C_t^{\text{energy}} + C_t^{\text{deg}} \right), 
\end{equation} 
where \(\lambda\in[0,1]\) controls the privacy-cost trade-off. 
The main privacy experiments use \(\lambda=1\), in which Eq.~\eqref{eq:total_reward} reduces to the privacy reward in Eq.~\eqref{eq:privacy_reward}. The cost model is detailed in Appendix~\ref{app:cost_model}.

\paragraph{Hierarchical signature mimicry policy.}

We use a signature mimicry mechanism based on a library \(\mathcal{S}=\{s_k\}_{k=1}^{K}\), where each signature \(s_k=[p^{(k)}_1,\ldots,p^{(k)}_{L_k}]\) is extracted from real appliance usage and \(L_k\) denotes its duration.
The hierarchical policy separates high-level signature selection from low-level battery execution.
At each decision point, the manager observes the state $s_t$, including the current load, time encoding, time-of-use price, battery state of charge, and remaining throughput budget, and outputs a discrete action
$a_t^{\mathrm{high}}\in\{0,1,\ldots,2K\}$.
Here, $a_t^{\mathrm{high}}=0$ denotes no mimicry; actions $1,\ldots,K$ replay the selected signature as a charging profile, while actions $K+1,\ldots,2K$ replay the corresponding signature as a discharging profile.
The low-level executor then replays the selected signature trajectory subject to battery feasibility constraints, including charge/discharge power limits, daily throughput budget, battery capacity, and the state-of-charge range. Detailed battery dynamics and constraint enforcement are provided in Appendix~\ref{app:constraints}.
If a replay trajectory violates these constraints, the executor terminates the action.
During replay, the manager remains inactive and resumes control only after the selected signature finishes or is stopped by battery constraints.



\paragraph{Policy training and deployment.}

Only the high-level manager is optimized in our settings.
The privacy probe is fixed, and the low-level executor deterministically replays selected signatures under battery constraints.
We train the manager policy \(\pi_\theta\) with PPO using the privacy reward in Eq.~\eqref{eq:privacy_reward}.
After training, only the learned manager policy, the signature library, and the deterministic battery executor are deployed.
The privacy probe is used only during offline training for reward computation and is not required during deployment.
Appliance-level labels are used only for constructing the signature library and for evaluation.
The deployed policy operates online using the observed aggregate load, battery state, time, electricity price, and remaining throughput budget, without access to the attacker's model.
The complete hierarchical signature mimicry procedure is summarized in Appendix~\ref{app:algorithm}.

\section{Experiments}
\label{sec:exp}


In this section, we evaluate whether a defender trained with a fixed privacy probe can generalize to heterogeneous black-box NILM attackers. We focus on three questions: whether the learned policy degrades unseen NILM models, whether the effect persists across datasets and test conditions, and which components of the proposed signature-mimicry mechanism contribute to the privacy gain.
For reproducibility, the implementation code of our work will be released upon acceptance.

\paragraph{Experimental setup.}
Below we present our experimental setting. 

\textit{Dataset.}
We conduct experiments on UK-DALE~\cite{kelly2015uk} and REDD~\cite{kolter2011redd} with consistent minute-level preprocessing and train--test protocols.
UK-DALE serves as the primary benchmark for black-box NILM evaluation.
We additionally repeat the defender policy, privacy probe training, and evaluation pipeline on REDD to test whether the proposed framework remains effective under a different dataset distribution.
Full dataset details are provided in Appendix~\ref{app:dataset}.

\textit{Signature library.}
We construct a trajectory-level signature library from real appliance activations, including 10 representative short-burst and medium-duration patterns from intermittent high-power appliances.
Details of signature extraction and library composition are provided in Appendix~\ref{app:signature}.
The library is constructed once from the UK-DALE House 1 training period and then fixed across all experiments, including the unseen-household and REDD evaluations; no appliance-level data from the target environments are used to adapt it.

\textit{Privacy probe.}
In the main experiments, we use a Transformer-encoder Seq2Seq (s2s) aggregate reconstruction model with a sequence prediction head as the privacy probe. Detailed model architecture and training process are presented in Appendix~\ref{app:privacy_probe}.
To verify that our framework is not dependent on a specific probe architecture, we further test CNN- and LSTM-based probes in Sec.~\ref{sec:ablation}, where all three probes yield comparable defense performance.

\textit{Evaluation attackers.}
We evaluate transferability against six independently trained NILM attackers. These attackers span five representative NILM paradigms:
optimization-based disaggregation (CO~\citep{hart1992nonintrusive}), probabilistic state-space models (FHMM~\citep{kolter2012approximate}),
convolutional sequence-to-point learning (S2P~\citep{zhang2018sequence}), denoising autoencoder-based NILM (DAE~\citep{kelly2015neural}),
and Transformer-based NILM (ELECTRIcity~\citep{sykiotis2022electricity} and BERT4NILM~\cite{yue2020bert4nilm}).
To ensure that the reported degradation reflects black-box transfer rather than overfitting to a specific inference architecture, all attacker models are trained independently following their original settings, with minor adjustments to match our sampling rate.

\textit{Defender training.}
The defender is implemented using PPO with Stable-Baselines3~\cite{stable-baselines3}.
Unless otherwise specified, the main experiments use \(\lambda=1\), corresponding to the pure privacy objective in Eq.~\eqref{eq:privacy_reward}.
Hyperparameters, observation features, battery settings, and training details are listed in Appendix~\ref{app:defender_setting}.

\textit{Evaluation metrics.}
Let $\pi_0$ denote the no-defense policy with $P_{\mathrm{batt}}(t)=0$.
To quantify privacy improvement during evaluation, we measure the increase in NILM inference risk caused by the defender relative to the unmasked baseline: $\Delta_{\mathrm{priv}}(f_\phi,\pi_\theta)=R(f_\phi,\pi_\theta)-R(f_\phi,\pi_0)$.
Positive values of $\Delta_{\mathrm{priv}}(f_\phi,\pi_\theta)$ indicate that the masked signal increases the attacker's inference error compared with the original aggregate signal.
In the experiments, we mainly report RMSE increase and F1 reduction because they respectively measure degradation in appliance-level power reconstruction and activation-state detection. MAE and SAE increases are reported as complementary error metrics.

\paragraph{Main results: Black-box transfer across NILM attackers.}
\label{sec:main_results}
\begin{wrapfigure}{r}{0.35\linewidth}
  \vspace{-8pt}
  \centering
  \includegraphics[
    width=\linewidth,
    trim=0 10 0 0,
    clip
  ]{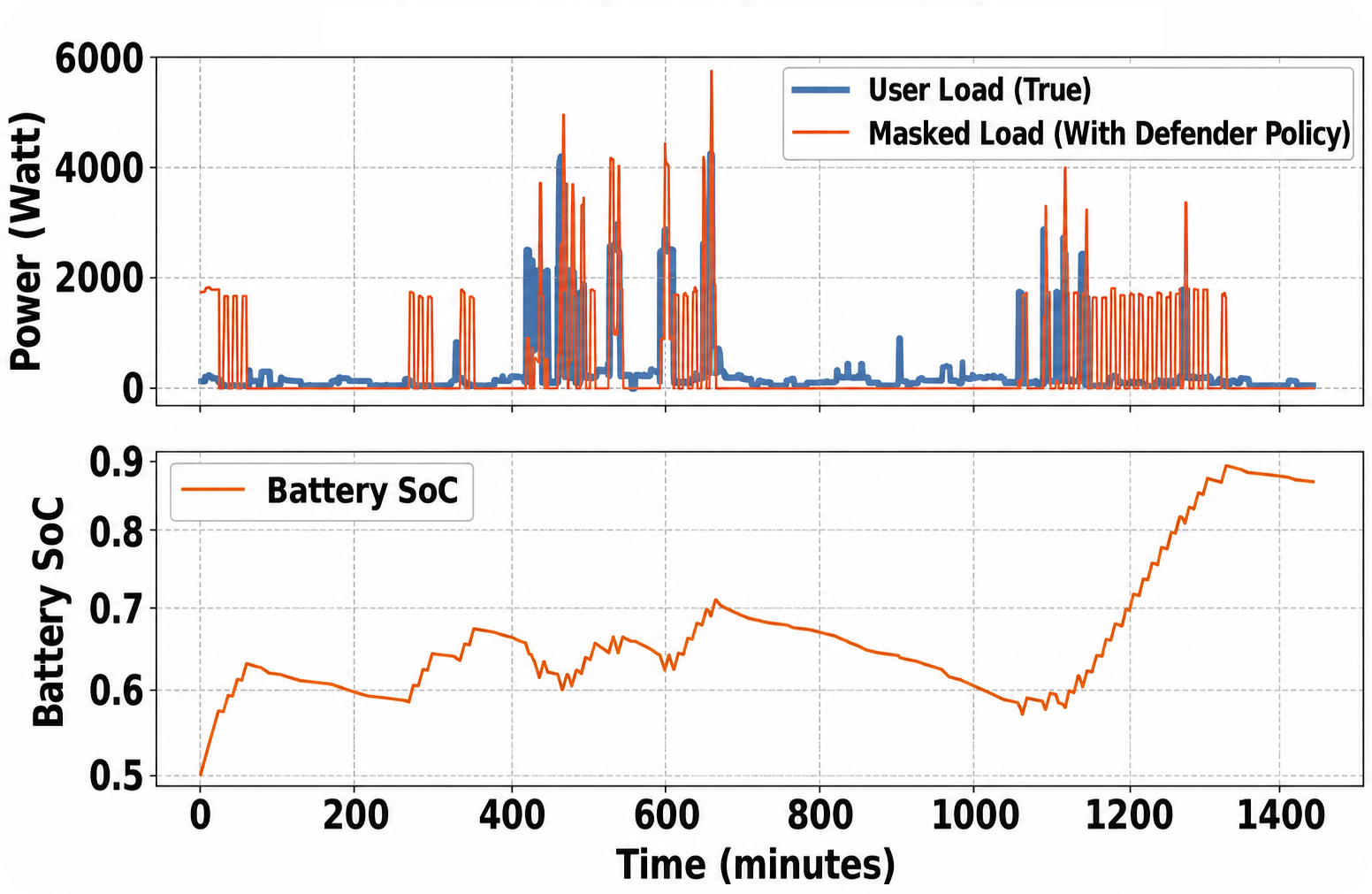}
  \vspace{-12pt}
  \caption{Illustration of the learned defender policy strategy and battery SoC on test day.}
  \label{fig:main_demo}
  \vspace{-15pt}
\end{wrapfigure}

\begin{figure}[t]
    \centering

    \includegraphics[width=0.99\linewidth]{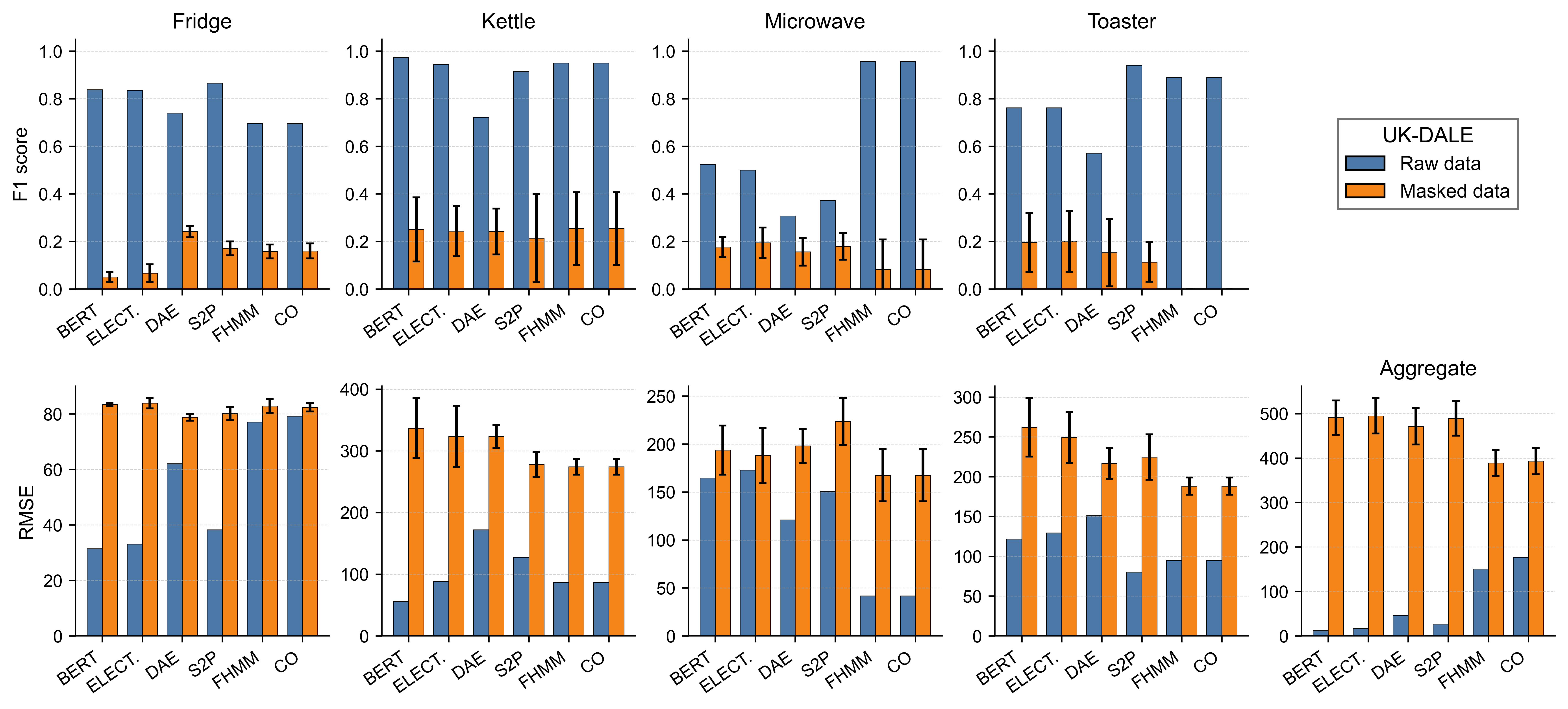}
    \vspace{-0.1em}

    \includegraphics[width=0.99\linewidth]{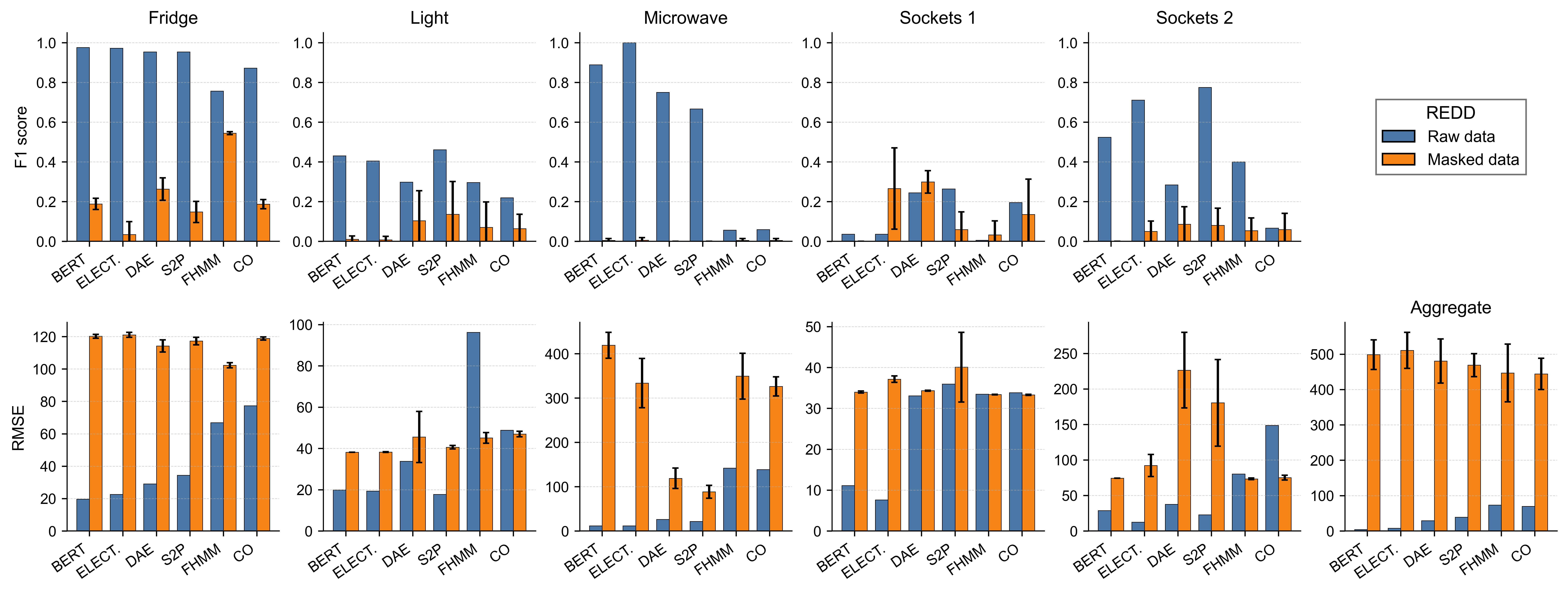}
    \vspace{-0.8em}

    \caption{
    Main black-box transfer results on UK-DALE and REDD, averaged over five random seeds.
    Top: UK-DALE results over four target appliances and the aggregate channel.
    Bottom: REDD results over five target appliances and the aggregate channel.
    The proposed defense consistently increases NILM inference error and reduces appliance detection performance across six unseen attackers.
    }
    \label{fig:main_results}
\end{figure}

\begin{table}[t]
\centering
\caption{
Overall black-box transfer performance averaged over five random seeds.
}
\label{tab:overall_transfer}
\resizebox{\linewidth}{!}{
\begin{tabular}{lccccccccc}
\toprule
\textbf{Dataset}
& \textbf{Cases}
& \textbf{RMSE Inc. $\uparrow$}
& \textbf{MAE Inc. $\uparrow$}
& \textbf{SAE Inc. $\uparrow$}
& \textbf{F1 Red. $\uparrow$}
& \textbf{RMSE Sig.}
& \textbf{MAE Sig.}
& \textbf{SAE Sig.}
& \textbf{F1 Sig.} \\
\midrule
UK-DALE
& 24
& 107.12\%
& 118.92\%
& 107.36\%
& 79.32\%
& 22/24
& 21/24
& 15/24
& \textbf{24/24} \\

REDD
& 30
& 165.90\%
& 127.35\%
& 196.33\%
& 80.10\%
& 22/30
& 22/30
& 25/30
& 25/30 \\
\bottomrule
\end{tabular}
}
\end{table}
Figure~\ref{fig:main_demo} illustrates that the learned defender selectively inserts realistic signature trajectories into the aggregate load while satisfying battery constraints.
We evaluate black-box transfer against six unseen NILM attackers over five random seeds.
RMSE and F1 are used as the primary privacy metrics, measuring appliance-level power reconstruction error and activation detection accuracy.

Figure~\ref{fig:main_results} reports attacker--appliance-level results on UK-DALE and REDD.
Across both datasets, the masked signals consistently increase NILM reconstruction errors and reduce appliance detection performance.
On UK-DALE, the defense substantially reduces F1 across all target appliances and attackers, while increasing RMSE for most attacker--appliance pairs.
The same trend holds on REDD, despite its different appliance composition and data distribution.
The only mixed cases are socket circuits, whose raw F1 scores are already low and whose composite plug-load patterns are less aligned with single-appliance signature mimicry.

Table~\ref{tab:overall_transfer} summarizes dataset-level relative changes, averaged over all attacker--appliance cases and seeds, excluding the aggregate channel.
RMSE Sig. and F1 Sig. ratio denotes the proportion of cases with statistically significant changes($p<0.05$).
The results show large average error increases and substantial F1 reductions on both datasets, indicating consistent degradation of unseen NILM attackers.
Overall, the REDD results reinforce our central claim that proxy-guided signature mimicry pipeline remains effective under a different dataset distribution and transfers to different black-box NILM settings.
Full numerical results are provided in Appendix~\ref{app:detailed_main_results}

\paragraph{Comparison with existing privacy defenses.}
\label{sec:baseline_comparison}
Table~\ref{tab:baseline_comparison} compares our method with random/rule-based battery baselines and two strong SM privacy defenses, DDQL-flat and DDQL-MI~\citep{shateri2021privacy,shateri2023privacy}.
Since existing work does not provide a directly comparable HRL and proxy-guided defense, these methods serve as the closest baselines with the same privacy-protection goal.
Our method achieves the strongest privacy protection, improving over DDQL-MI by 13.61 percentage points in RMSE increase and 6.65 percentage points in F1 reduction.
It also yields more significant degradation cases, reaching 22/24 for RMSE and 24/24 for F1.
Meanwhile, our method has the lowest energy and total cost, and requires less training time than DDQL-based baselines because privacy rewards are computed only when a mimic signature is inserted.
The negative energy cost is due to time-of-use arbitrage, where the battery charges during low-price periods and discharges during high-price periods.

\begin{table}[t]
\centering
\caption{
Comparison with existing privacy defenses on UK-DALE.
}
\label{tab:baseline_comparison}
\resizebox{\linewidth}{!}{
\begin{tabular}{lcccccccc}
\toprule
\textbf{Method}
& \textbf{RMSE Inc. $\uparrow$}
& \textbf{F1 Red. $\uparrow$}
& \textbf{RMSE Sig.}
& \textbf{F1 Sig.}
& \textbf{Energy Cost}
& \textbf{Deg. Cost}
& \textbf{Total Cost}
& \textbf{Train Time} \\
\midrule

No defense
& 0.00\%
& 0.00\%
& --
& --
& 0.000
& 0.000
& 0.000
& -- \\

Random baseline
& 18.58\%
& 7.25\%
& 7/24
& 8/24
& 0.036
& \textbf{0.029}
& 0.065
& -- \\

Rule-based baseline
& 32.44\%
& 24.56\%
& 10/24
& 10/24
& 0.177
& 0.207
& 0.384
& -- \\

DDQL-flat
& 74.03\%
& 70.06\%
& 19/24
& 20/24
& -0.686
& 0.456
& -0.230
& 3.2h \\

DDQL-MI
& 93.51\%
& 72.67\%
& 15/24
& 20/24
& -0.482
& 0.600
& 0.118
& 6.6h \\

\textbf{Ours}
& \textbf{107.12\%}
& \textbf{79.32\%}
& \textbf{22/24}
& \textbf{24/24}
& \textbf{-1.224}
& 0.460
& \textbf{-0.764}
& 1.2h \\

\bottomrule
\end{tabular}
}
\end{table}

\begin{wrapfigure}{r}{0.35\linewidth}
\vspace{-0.6em}
\centering
\includegraphics[width=\linewidth]{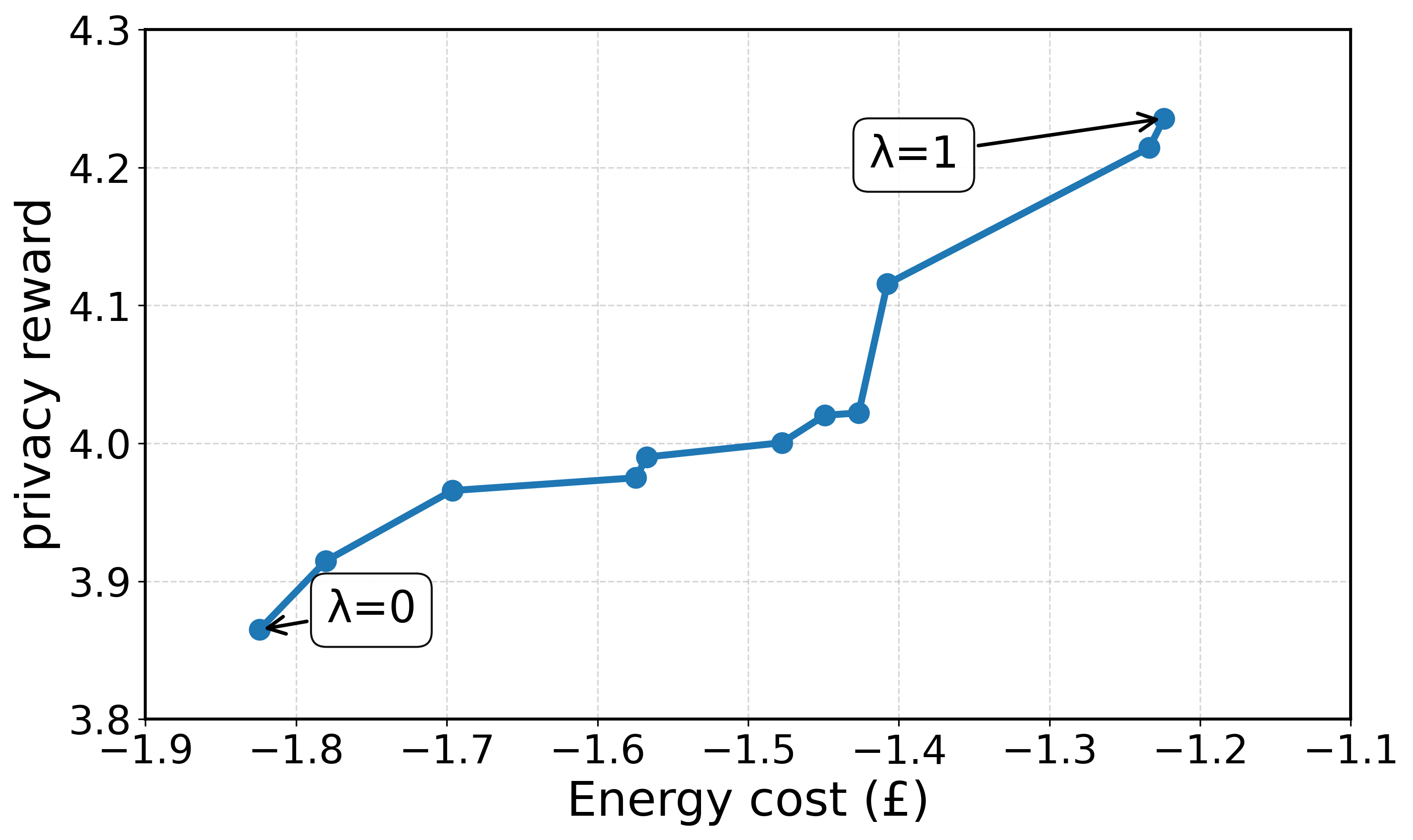}
\vspace{-0.8em}
\caption{
Pareto frontier of privacy reward versus energy cost across the $\lambda$ sweep.
}
\label{fig:lambda_pareto}
\vspace{-0.8em}
\end{wrapfigure}

\paragraph{Ablation studies.}
\label{sec:ablation}
We conduct ablation studies to examine how the privacy probe architecture and signature library size affect black-box transfer performance.
Table~\ref{tab:probe_ablation} shows that Seq2Seq, LSTM, and CNN probes all yield strong black-box transfer, with RMSE increases above 94\% and F1 reductions above 77\%.
Seq2Seq performs best overall, suggesting that a more expressive surrogate probe provides a stronger learning signal, but the defense is not tied to a specific probe architecture.
Table~\ref{tab:signature_size_ablation} shows that reducing the library generally weakens privacy protection, with RMSE increase dropping from 107.12\% using 10 signatures to 66.45\% using 3 signatures.
However, the 9-signature setting remains close to the full library, indicating that moderate pruning has limited impact while richer libraries improve mimicry diversity.

\begin{table}[t]
\centering

\begin{minipage}[t]{0.45\linewidth}
\centering
\caption{Effect of privacy probe architecture on UK-DALE.}
\label{tab:probe_ablation}
\resizebox{\linewidth}{!}{
\begin{tabular}{lcccc}
\toprule
\textbf{Probe}
& \textbf{RMSE Inc.}
& \textbf{F1 Red.}
& \textbf{RMSE Sig.}
& \textbf{F1 Sig.} \\
\midrule
Seq2Seq & \textbf{107.12\%} & \textbf{79.32\%} & \textbf{22/24} & \textbf{24/24} \\
LSTM    & 100.71\%          & 78.75\%          & \textbf{22/24} & 22/24 \\
CNN     & 94.13\%           & 77.46\%          & 21/24          & 21/24 \\
\bottomrule
\end{tabular}
}
\end{minipage}
\hfill
\begin{minipage}[t]{0.45\linewidth}
\centering
\caption{Effect of signature library size on UK-DALE.}
\label{tab:signature_size_ablation}
\resizebox{\linewidth}{!}{
\begin{tabular}{lcccc}
\toprule
\textbf{Library Size}
& \textbf{RMSE Inc.}
& \textbf{F1 Red.}
& \textbf{RMSE Sig.}
& \textbf{F1 Sig.} \\
\midrule
10 (main) & \textbf{107.12\%} & \textbf{79.32\%} & \textbf{22/24} & \textbf{24/24} \\
9         & 103.82\%          & 77.54\%          & 21/24          & 22/24 \\
7         & 93.95\%           & 61.87\%          & 19/24          & 20/24 \\
5         & 85.03\%           & 68.65\%          & 18/24          & 19/24 \\
3         & 66.45\%           & 53.54\%          & 17/24          & 17/24 \\
\bottomrule
\end{tabular}
}
\end{minipage}
\end{table}

\paragraph{Robustness and practical constraints.}
\label{sec:robustness_constraints}
Below we provide analysis about our defender robustness and physical configuration sensitivity.

\textit{Battery configuration sensitivity.}
\begin{table}[t]
\centering
\caption{
Battery configuration sensitivity on UK-DALE.
}
\label{tab:battery_sensitivity}
\resizebox{0.9\linewidth}{!}{
\begin{tabular}{lccc ccccc}
\toprule
\multirow{2}{*}{\textbf{Impact Factor}}
& \multicolumn{3}{c}{\textbf{Battery Config}}
& \multirow{2}{*}{\textbf{RMSE Inc.} $\uparrow$}
& \multirow{2}{*}{\textbf{F1 Red.} $\uparrow$}
& \multirow{2}{*}{\textbf{Agg. $\Delta$RMSE} $\uparrow$}
& \multirow{2}{*}{\textbf{RMSE Sig.}}
& \multirow{2}{*}{\textbf{F1 Sig.}} \\
\cmidrule(lr){2-4}
& $\mathbf{C_{\max}}$ \textbf{(kWh)}
& $\mathbf{P_{\max}}$ \textbf{(kW)}
& $\mathbf{T_{\max}}$ \textbf{(kWh)}
& & & & & \\
\midrule

\textbf{Main setting}
& \textbf{8} & \textbf{4} & \textbf{8}
& \textbf{107.1\%}
& \textbf{79.3\%}
& \textbf{413.9}
& \textbf{22/24}
& \textbf{24/24} \\

\midrule

\multirow{3}{*}{Battery capacity}
& 2  & 4 & 8
& 56.0\%
& 32.6\%
& 239.2
& 12/24
& 14/24 \\

& 4  & 4 & 8
& 60.2\%
& 42.2\%
& 315.9
& 17/24
& 20/24 \\

& 12 & 4 & 8
& 107.2\%
& 82.6\%
& 405.4
& 22/24
& 24/24 \\

\midrule

\multirow{3}{*}{Power limit}
& 8 & 1 & 8
& 85.1\%
& 62.3\%
& 247.9
& 13/24
& 14/24 \\

& 8 & 2 & 8
& 110.2\%
& 79.3\%
& 393.3
& 21/24
& 24/24 \\

& 8 & 6 & 8
& 98.0\%
& 80.9\%
& 392.6
& 22/24
& 24/24 \\

\midrule

\multirow{3}{*}{Throughput budget}
& 8 & 4 & 2
& 35.7\%
& 24.4\%
& 176.0
& 11/24
& 11/24 \\

& 8 & 4 & 4
& 72.2\%
& 59.0\%
& 288.1
& 15/24
& 18/24 \\

& 8 & 4 & 6
& 90.2\%
& 76.1\%
& 411.1
& 21/24
& 23/24 \\

\bottomrule
\end{tabular}
}
\end{table}
Table~\ref{tab:battery_sensitivity} studies the effect of battery capacity $C_{\max}$, power limit $P_{\max}$, and daily throughput budget $T_{\max}$, averaged over six unseen NILM attackers and four target appliances.
Privacy protection improves substantially when capacity and throughput increase, but shows diminishing returns beyond 8 kWh capacity and 2 kW power.
Among the three factors, daily throughput budget has the strongest influence, since it directly limits how much mimic-signature replay can be sustained within a day.

\textit{Privacy-cost trade-off.}
Figure~\ref{fig:lambda_pareto} shows that the privacy-cost trade-off factor $\lambda$ provides a smooth control knob between privacy and energy cost.
The $\lambda=0$ policy minimizes energy cost through time-of-use arbitrage, whereas $\lambda=1$ maximizes privacy reward.

\textit{Multi-day and multi-household robustness.}
Table~\ref{tab:heldout_day_generalization} shows that the same defender generalizes without retraining across seasons, weekdays/weekends, and households.
All test days show increased RMSE and reduced F1, with F1 reduction ranging from 32.1\% to 93.7\%.
The effect size varies with appliance activity and raw NILM difficulty, so we report both raw and masked metrics rather than relying only on relative changes to avoid overstating gains on low-error or low-activity days.

\begin{table}[t]
\centering
\caption{
Generalization across the main and UK-DALE test days.
}
\label{tab:heldout_day_generalization}
\resizebox{\linewidth}{!}{
\begin{tabular}{lccccccc}
\toprule
\textbf{Test Day}
& \textbf{Active App.}
& \textbf{RMSE Raw}
& \textbf{RMSE Masked}
& \textbf{RMSE Inc. $\uparrow$}
& \textbf{F1 Raw}
& \textbf{F1 Masked}
& \textbf{F1 Red. $\downarrow$} \\
\midrule

\textbf{2013-03-18 H1 (Spring-Monday, Main)}
& \textbf{4}
& \textbf{95.95}
& \textbf{197.27}
& \textbf{105.6\%}
& \textbf{0.773}
& \textbf{0.161}
& \textbf{79.2\%} \\

\midrule

2013-01-18 H1 (Winter-Friday)
& 4
& 113.01
& 176.38
& 56.1\%
& 0.687
& 0.361
& 47.5\% \\

2013-03-30 H1 (Spring-Saturday)
& 1
& 7.39
& 89.76
& 1115.4\%
& 0.225
& 0.014
& 93.6\% \\

2013-03-30 H4 (Spring-Saturday)
& 2
& 64.36
& 135.08
& 109.9\%
& 0.837
& 0.568
& 32.1\% \\

2013-07-15 H1 (Summer-Monday)
& 4
& 72.62
& 150.57
& 107.3\%
& 0.698
& 0.196
& 71.9\% \\

2013-07-15 H2 (Summer-Monday)
& 3
& 58.85
& 164.06
& 178.8\%
& 0.643
& 0.040
& 93.7\% \\

\bottomrule
\end{tabular}
}
\end{table}
\begin{table}[t]
\centering
\caption{Generalization across the main and REDD test days.}
\label{tab:heldout_day_generalization_redd}
\resizebox{\columnwidth}{!}{
\begin{tabular}{lccccccc}
\toprule
\textbf{Test Day} &
\textbf{Active App.} &
\textbf{RMSE Raw} &
\textbf{RMSE Masked} &
\textbf{RMSE Inc. $\uparrow$} &
\textbf{F1 Raw} &
\textbf{F1 Masked} &
\textbf{F1 Red. $\downarrow$} \\
\midrule

\textbf{2011-04-29 H2 (Spring-Friday, Main)} &
\textbf{5} &
\textbf{44.48} &
\textbf{115.47} &
\textbf{165.9\%} &
\textbf{0.493} &
\textbf{0.098} &
\textbf{80.1\%} \\
\midrule
2011-04-30 H2 (Spring-Saturday) & 5 & 61.53 & 123.75 & 101.1\% & 0.545 & 0.125 & 77.1\% \\
2011-04-26 H1 (Spring-Tuesday) & 3 & 80.36 & 170.30 & 111.9\% & 0.698 & 0.308 & 55.9\% \\
2011-05-25 H3 (Spring-Wednesday) & 3 & 111.89 & 173.51 & 55.1\% & 0.580 & 0.336 & 42.2\% \\
2011-05-01 H4 (Spring-Sunday) & 1 & 57.36 & 71.04 & 23.8\% & 0.934 & 0.360 & 61.4\% \\

\bottomrule
\end{tabular}
}
\end{table}

\section{Conclusion}
\label{sec:conclusion}
We studied black-box NILM privacy defense under attacker uncertainty, and proposed a proxy-guided HRL framework in which a fixed self-supervised aggregate-structure probe provides a surrogate training signal, while a curated signature library defines a physically realizable, task-aligned perturbation class. Experiments on UK-DALE and REDD show that the learned defender transfers to six unseen NILM attackers spanning optimization-based, probabilistic, and neural sequence architectures, and remains effective across households, 
seasons, and weekday/weekend conditions without retraining. These results suggest a more general design principle for black-box inference defense: pairing a task-aligned surrogate signal with a structured perturbation class can yield strong cross-attacker transfer without access to the deployed inference model.

\section{Limitations and Future Work}
Three limitations point to natural extensions.
\emph{First}, the curated signature library has finite coverage and is most effective for intermittent high-power appliances; composite low-power loads such as socket circuits on REDD remain harder to mask via single-signature mimicry. Learning the signature library jointly with the policy, or extending it to multi-appliance composite trajectories, is a promising direction.

\emph{Second}, our theoretical analysis (Appendix~\ref{app:theory_analysis}) establishes a local sufficient condition for proxy-guided improvement of the expected attacker risk; characterizing global behavior across attacker families, or quantifying the surrogate–target gap under specific NILM architectures, remains open.

\emph{Third}, evaluation is conducted offline against trained but fixed NILM attackers; deployment in a live SM environment would additionally require handling forecasting uncertainty in user load, online battery state estimation, and adaptive attackers that 
bserve the masked signal over time. Stress-testing the framework against adversarially adaptive NILM attackers is, in particular, an important direction for future work.

\newpage

\bibliographystyle{plain}
\bibliography{example_paper}

\newpage
\appendix
\textbf{Appendix}


\section{Cost Modeling and Battery Constraints}
\label{app:cost}

This appendix provides the battery feasibility constraints and operational cost model used in this work.
The feasibility constraints define the executable action space of the low-level battery executor, while the cost model quantifies the economic and degradation implications of battery operation.
Unless otherwise specified, the main privacy experiments use \(\lambda=1\), corresponding to privacy-focused optimization, and report electricity and battery degradation costs separately.
\subsection{Cost Modelling: Battery Degradation and Operational Cost}
\label{app:cost_model}
Battery operation incurs both time-of-use electricity cost and battery degradation cost.
These quantities correspond to \(C_t^{\mathrm{energy}}\) and \(C_t^{\mathrm{deg}}\) in Eq.~\eqref{eq:total_reward}.
In the main privacy experiments, \(\lambda=1\), so Eq.~\eqref{eq:total_reward} reduces to the privacy reward in Eq.~\eqref{eq:privacy_reward}; the following costs are evaluated and reported separately.

We adopt a time-of-use electricity tariff \(c_{\mathrm{e}}(t)\).
Under the sign convention \(P_{\text{batt}}(t)>0\) for charging and \(P_{\text{batt}}(t)<0\) for discharging, the ToU energy cost is defined as
\begin{equation}
\label{eq:app_tou_cost}
C_t^{\mathrm{energy}}
=
c_{\mathrm{e}}(t)\,\Delta t
\begin{cases}
\dfrac{P_{\text{batt}}(t)}{\eta_{\mathrm{c}}}, 
& P_{\text{batt}}(t) > 0 \quad \text{(charging)}, \\[4pt]
\eta_{\mathrm{d}}\, P_{\text{batt}}(t),
& P_{\text{batt}}(t) < 0 \quad \text{(discharging)}, \\[6pt]
0, 
& P_{\text{batt}}(t) = 0 .
\end{cases}
\end{equation}
Here, \(C_t^{\mathrm{energy}}\) may take negative values, reflecting cost savings when the battery discharges to offset grid consumption during high-price periods.
This corresponds to temporal energy shifting rather than feed-in tariffs, as no electricity is sold back to the grid.

For simplicity and clarity of analysis, we set \(\eta_{\mathrm{c}}=\eta_{\mathrm{d}}=1\) in all experiments.
Under this setting, Eq.~\eqref{eq:app_tou_cost} simplifies to
\begin{equation}
\label{eq:app_tou_cost_simplified}
C_t^{\mathrm{energy}}
=
c_{\mathrm{e}}(t)\,\Delta t\,P_{\text{batt}}(t).
\end{equation}

Battery cycling also induces long-term degradation.
Following throughput-based cycle aging models, the degradation cost is modeled as proportional to the processed battery energy:
\begin{equation}
\label{eq:app_degradation_cost}
    C_t^{\mathrm{deg}}
    =
    c_{\mathrm{deg}}
    \left|
        P_{\text{batt}}(t)
    \right|
    \Delta t,
\end{equation}
where \(c_{\mathrm{deg}}\) denotes the amortized degradation cost per processed kilowatt-hour.

The overall operational cost at time \(t\) is therefore
\begin{equation}
\label{eq:app_operational_cost}
    C_t^{\mathrm{op}}
    =
    C_t^{\mathrm{energy}}
    +
    C_t^{\mathrm{deg}}.
\end{equation}
When operational costs are considered during policy optimization, \(C_t^{\mathrm{op}}\) is combined with the privacy reward through the trade-off coefficient \(\lambda\) in Eq.~\eqref{eq:total_reward}.

\subsection{Battery Feasibility Constraints}
\label{app:constraints}
The masked load follows the sign convention: \(P_{\text{batt}}(t)>0\) denotes battery charging and increases the reported aggregate load, while \(P_{\text{batt}}(t)<0\) denotes battery discharging and decreases the reported aggregate load.

All battery actions generated by the low-level executor are subject to hard physical constraints.
First, the battery power is bounded by the maximum charging and discharging rates:
\begin{equation}
\label{eq:app_power_constraint}
    -P_{\max}^{\mathrm{dis}}
    \leq
    P_{\text{batt}}(t)
    \leq
    P_{\max}^{\mathrm{ch}},
\end{equation}
where \(P_{\max}^{\mathrm{ch}}\) and \(P_{\max}^{\mathrm{dis}}\) denote the maximum charging and discharging power, respectively.

The battery state of charge evolves according to
\begin{equation}
\label{eq:app_soc_dynamics}
    \mathrm{SoC}_{t+1}
    =
    \mathrm{SoC}_{t}
    +
    \frac{\eta_{\mathrm{c}} [P_{\text{batt}}(t)]_+ \Delta t}{E_{\max}}
    -
    \frac{[-P_{\text{batt}}(t)]_+ \Delta t}{\eta_{\mathrm{d}} E_{\max}},
\end{equation}
where \(E_{\max}\) is the battery capacity, \(\eta_{\mathrm{c}}\) and \(\eta_{\mathrm{d}}\) are the charging and discharging efficiencies, \(\Delta t\) is the sampling interval, and \([x]_+=\max(x,0)\).

To avoid unrealistic operation near full depletion or full charge, the battery is constrained to operate within a safe state-of-charge range:
\begin{equation}
\label{eq:app_soc_bounds}
    0.1
    \leq
    \mathrm{SoC}_{t}
    \leq
    0.9.
\end{equation}

When a daily throughput budget is used, the cumulative processed battery energy within one day is additionally constrained by
\begin{equation}
\label{eq:app_throughput_budget}
    \sum_{t \in \mathcal{D}}
    \left|
        P_{\text{batt}}(t)
    \right|
    \Delta t
    \leq
    B_{\mathrm{day}},
\end{equation}
where \(\mathcal{D}\) denotes the set of timesteps within the day and \(B_{\mathrm{day}}\) is the daily throughput budget.

During signature replay, a high-level action first defines a nominal battery trajectory:
\begin{equation}
\label{eq:app_nominal_replay}
    \tilde{P}_{\text{batt}}(t+\tau)
    =
    \sigma(a_t^{\mathrm{high}})
    p^{(k)}_{\tau},
    \qquad
    \tau=1,\ldots,L_k,
\end{equation}
where \(\sigma(a_t^{\mathrm{high}})=+1\) denotes charging and \(\sigma(a_t^{\mathrm{high}})=-1\) denotes discharging.
The actually executed battery action \(P_{\text{batt}}(t+\tau)\) is obtained by enforcing the constraints in Eqs.~\eqref{eq:app_power_constraint}--\eqref{eq:app_throughput_budget}.
If the nominal replay violates the charge/discharge power limits, the state-of-charge range, or the remaining throughput budget, the executor clips the action to the feasible range or terminates the replay.

\section{Theoretical Analysis for Proxy-Guided Signature Manipulation}
\label{app:theory_analysis}

This appendix provides a sufficient-condition analysis for the proxy-guided objective used in Section~\ref{sec:theoretical_rationale}.
The goal is not to establish a universal guarantee for transfer to all possible NILM attackers.
Rather, the analysis formalizes a local condition under which optimizing a surrogate privacy probe can improve the expected inference risk over a heterogeneous attacker family.
This condition can be viewed as a policy-constrained analogue of surrogate-target alignment in transfer-based black-box adversarial attacks.

\subsection{Notation and Objective}

Let $\mathcal{D}$ denote the data distribution over aggregate load windows and appliance-level targets.
For a defense policy $\pi_\theta$, let $\mathbf{x}^{\pi_\theta}$ denote the masked aggregate window induced by feasible battery actions.
For a NILM attacker $f_\phi$, the inference risk is
\begin{equation}
\label{eq:app_attacker_risk}
    R(f_\phi,\pi_\theta)
    =
    \mathbb{E}_{(\mathbf{x},\mathbf{y})\sim\mathcal{D}}
    \left[
        \mathcal{L}
        \left(
            f_\phi(\mathbf{x}^{\pi_\theta}),
            \mathbf{y}
        \right)
    \right],
\end{equation}
where $\mathcal{L}$ measures appliance-level reconstruction error.
The ideal robust objective under attacker uncertainty is
\begin{equation}
\label{eq:app_avg_risk}
    \mathcal{R}_{\mathrm{avg}}(\pi_\theta)
    =
    \mathbb{E}_{f_\phi\sim\mu}
    \left[
        R(f_\phi,\pi_\theta)
    \right],
\end{equation}
where $\mu$ denotes the latent distribution over possible NILM attackers.
Since $\mu$ is unknown during training, the defender instead optimizes the proxy risk induced by the fixed aggregate-structure privacy probe:
\begin{equation}
\label{eq:app_proxy_risk}
    R_p(\pi_\theta)
    =
    R(f_\eta,\pi_\theta).
\end{equation}

Let
\begin{equation}
\label{eq:app_representation}
z_\theta = \psi(\mathbf{x}^{\pi_\theta})
\end{equation}
denote the NILM-relevant representation induced by the current policy.
The representation map $\psi(\cdot)$ is used only as an analytical abstraction of task-level commonality among NILM attackers, rather than as an explicitly learned feature extractor shared by all models.

For compactness, we write the representation-level proxy risk and expected attacker-family risk as
\begin{equation}
\label{eq:app_representation_risks}
    r_p(z_\theta)
    =
    R_p(\pi_\theta),
    \qquad
    r_{\mathrm{avg}}(z_\theta)
    =
    \mathcal{R}_{\mathrm{avg}}(\pi_\theta),
\end{equation}
where the dependence on $\mathcal{D}$ is implicit.
The corresponding representation-level risk gradients are
\begin{equation}
\label{eq:app_representation_gradients}
    u_p
    =
    \nabla_z r_p(z_\theta),
    \qquad
    u_{\mathrm{avg}}
    =
    \nabla_z r_{\mathrm{avg}}(z_\theta).
\end{equation}

\subsection{Policy-Reachable Representation Directions}

The defense policy cannot manipulate arbitrary directions in the representation space.
It can only induce changes through feasible battery actions, subject to capacity, power, state-of-charge, throughput, and cost constraints.
Therefore, the relevant transfer condition should be stated only for directions that are locally reachable by the policy.

Let
\begin{equation}
\label{eq:app_jacobian}
    J_\theta
    =
    \frac{\partial z_\theta}{\partial \theta}
\end{equation}
denote the local policy-to-representation Jacobian.
The local policy-reachable representation subspace is
\begin{equation}
\label{eq:app_reachable_subspace}
    \mathcal{T}_{\theta}
    =
    \mathrm{Range}(J_\theta).
\end{equation}
Let $P_\theta$ denote the orthogonal projection onto $\mathcal{T}_{\theta}$.
For any representation-level gradient $u$, define its reachable component as
\begin{equation}
\label{eq:app_reachable_component}
    u^{\mathrm{reach}}
    =
    P_\theta u.
\end{equation}
Since directions orthogonal to $\mathcal{T}_{\theta}$ cannot be induced by local policy changes, they do not affect the policy-gradient update:
\begin{equation}
\label{eq:app_projection_identity}
    J_\theta^\top u
    =
    J_\theta^\top u^{\mathrm{reach}}.
\end{equation}

\subsection{Local Surrogate-Target Alignment}

We now state a local sufficient condition connecting the proxy risk and the expected attacker-family risk.
The condition is defined in the metric induced by the policy-to-representation mapping.

Let
\begin{equation}
\label{eq:app_policy_metric}
    M_\theta
    =
    J_\theta J_\theta^\top
\end{equation}
denote the policy-induced metric on the reachable representation subspace.
For any two reachable representation directions $a,b\in\mathcal{T}_{\theta}$, their corresponding policy-gradient inner product is
\begin{equation}
\label{eq:app_metric_identity}
    \left\langle
        J_\theta^\top a,
        J_\theta^\top b
    \right\rangle
    =
    a^\top
    M_\theta
    b.
\end{equation}

\paragraph{Condition A1: Policy-induced structure-to-risk alignment.}
At the current policy parameter $\theta$, suppose the reachable proxy-risk direction and the reachable expected-risk direction satisfy
\begin{equation}
\label{eq:app_metric_alignment_condition}
    \left(
        u_p^{\mathrm{reach}}
    \right)^\top
    M_\theta
    u_{\mathrm{avg}}^{\mathrm{reach}}
    \geq
    \gamma
    \left(
        u_p^{\mathrm{reach}}
    \right)^\top
    M_\theta
    u_p^{\mathrm{reach}}
    -
    \varepsilon_{\mathrm{tr}},
    \qquad
    \gamma>0,
\end{equation}
where $\varepsilon_{\mathrm{tr}}\geq0$ captures residual transfer error.

This condition states that, after accounting for how feasible signature-mimicry actions change aggregate structures, the probe-risk direction is locally aligned with the expected NILM-risk direction over the attacker family.
It does not require the probe to predict appliance-level outputs or replicate any deployed NILM attacker.
Instead, it assumes that, within the signature-reachable action space, aggregate structures that strongly degrade probe reconstruction are also structures that affect downstream NILM disaggregation.

Because the attacker distribution $\mu$ is unknown, this condition cannot be directly verified during training.
Its practical relevance is evaluated empirically by testing whether policies trained against $f_\eta$ degrade unseen NILM attackers.

\subsection{Policy-Level Alignment}

The following proposition shows that Condition A1 directly induces alignment between the proxy-gradient direction and the expected-risk gradient in policy-parameter space.

\paragraph{Proposition 1: Policy-level proxy alignment.}
Under Condition A1,
\begin{equation}
\label{eq:app_policy_alignment_metric}
    \left\langle
        \nabla_\theta R_p(\pi_\theta),
        \nabla_\theta \mathcal{R}_{\mathrm{avg}}(\pi_\theta)
    \right\rangle
    \geq
    \gamma
    \left\|
        \nabla_\theta R_p(\pi_\theta)
    \right\|^2
    -
    \varepsilon_{\mathrm{tr}}.
\end{equation}

\paragraph{Proof.}
By the chain rule and Eq.~\eqref{eq:app_projection_identity},
\begin{equation}
\label{eq:app_chain_rule_policy}
    \nabla_\theta R_p(\pi_\theta)
    =
    J_\theta^\top u_p^{\mathrm{reach}},
    \qquad
    \nabla_\theta \mathcal{R}_{\mathrm{avg}}(\pi_\theta)
    =
    J_\theta^\top u_{\mathrm{avg}}^{\mathrm{reach}}.
\end{equation}
Therefore,
\begin{align}
    \left\langle
        \nabla_\theta R_p(\pi_\theta),
        \nabla_\theta \mathcal{R}_{\mathrm{avg}}(\pi_\theta)
    \right\rangle
    &=
    \left\langle
        J_\theta^\top u_p^{\mathrm{reach}},
        J_\theta^\top u_{\mathrm{avg}}^{\mathrm{reach}}
    \right\rangle
    \nonumber\\
    &=
    \left(
        u_p^{\mathrm{reach}}
    \right)^\top
    J_\theta J_\theta^\top
    u_{\mathrm{avg}}^{\mathrm{reach}}
    \nonumber\\
    &=
    \left(
        u_p^{\mathrm{reach}}
    \right)^\top
    M_\theta
    u_{\mathrm{avg}}^{\mathrm{reach}}.
\end{align}
Applying Condition A1 gives
\begin{align}
    \left\langle
        \nabla_\theta R_p(\pi_\theta),
        \nabla_\theta \mathcal{R}_{\mathrm{avg}}(\pi_\theta)
    \right\rangle
    &\geq
    \gamma
    \left(
        u_p^{\mathrm{reach}}
    \right)^\top
    M_\theta
    u_p^{\mathrm{reach}}
    -
    \varepsilon_{\mathrm{tr}}
    \nonumber\\
    &=
    \gamma
    \left\|
        J_\theta^\top
        u_p^{\mathrm{reach}}
    \right\|^2
    -
    \varepsilon_{\mathrm{tr}}
    \nonumber\\
    &=
    \gamma
    \left\|
        \nabla_\theta R_p(\pi_\theta)
    \right\|^2
    -
    \varepsilon_{\mathrm{tr}}.
\end{align}
This proves Eq.~\eqref{eq:app_policy_alignment_metric}.
\hfill$\square$

\subsection{Local Improvement of the Robust Objective}

The previous proposition establishes policy-level alignment.
We next state a local improvement result for a small proxy-gradient ascent step.

\paragraph{Condition A2: Local smoothness.}
Assume that $\mathcal{R}_{\mathrm{avg}}(\pi_\theta)$ is locally $L$-smooth as a function of $\theta$.
That is, for a sufficiently small update $\Delta\theta$,
\begin{equation}
\label{eq:app_smoothness}
    \mathcal{R}_{\mathrm{avg}}(\pi_{\theta+\Delta\theta})
    \geq
    \mathcal{R}_{\mathrm{avg}}(\pi_\theta)
    +
    \left\langle
        \nabla_\theta
        \mathcal{R}_{\mathrm{avg}}(\pi_\theta),
        \Delta\theta
    \right\rangle
    -
    \frac{L}{2}
    \|\Delta\theta\|^2.
\end{equation}

\paragraph{Proposition 2: Proxy-guided local improvement.}
Let the policy be updated by a small proxy-gradient ascent step:
\begin{equation}
\label{eq:app_proxy_update}
    \theta^+
    =
    \theta
    +
    \eta
    \nabla_\theta R_p(\pi_\theta),
\end{equation}
where $\eta>0$ is the step size.
Under Conditions A1 and A2,
\begin{align}
\label{eq:app_local_improvement}
    \mathcal{R}_{\mathrm{avg}}(\pi_{\theta^+})
    -
    \mathcal{R}_{\mathrm{avg}}(\pi_\theta)
    \geq
    \eta
    \left[
        \gamma
        \left\|
            \nabla_\theta R_p(\pi_\theta)
        \right\|^2
        -
        \varepsilon_{\mathrm{tr}}
    \right]
    -
    \frac{L\eta^2}{2}
    \left\|
        \nabla_\theta R_p(\pi_\theta)
    \right\|^2.
\end{align}
Thus, for sufficiently small $\eta$, the proxy-guided update improves the expected attacker-family risk whenever the surrogate-target alignment term dominates the residual transfer error.

\paragraph{Proof.}
Set
\begin{equation}
    \Delta\theta
    =
    \eta
    \nabla_\theta R_p(\pi_\theta).
\end{equation}
Substituting this update into Eq.~\eqref{eq:app_smoothness} gives
\begin{align}
    \mathcal{R}_{\mathrm{avg}}(\pi_{\theta^+})
    -
    \mathcal{R}_{\mathrm{avg}}(\pi_\theta)
    &\geq
    \eta
    \left\langle
        \nabla_\theta
        \mathcal{R}_{\mathrm{avg}}(\pi_\theta),
        \nabla_\theta R_p(\pi_\theta)
    \right\rangle
    -
    \frac{L\eta^2}{2}
    \left\|
        \nabla_\theta R_p(\pi_\theta)
    \right\|^2.
\end{align}
Applying Proposition~1 yields Eq.~\eqref{eq:app_local_improvement}.
\hfill$\square$

\subsection{Empirical evidence of Condition A1.}
Although Condition A1 cannot be directly verified at training time, its empirical implication 
is testable: a policy trained against $f_\eta$ should degrade unseen NILM attackers in $\mathcal{F}$. 
The cross-attacker transfer results in Section~\ref{sec:main_results} (Table~\ref{tab:overall_transfer}) 
provide consistent evidence: across 24 attacker–appliance pairs on UK-DALE and 30 on REDD, 
the proxy-guided policy increases attacker reconstruction error by over 100\% on average and reduces 
F1 by ~80\%. This is the empirical signature of $\gamma > 0$ holding broadly across the attacker family.

\subsection{Signature Mimic as a Structured Perturbation Class}

The main text motivates signature mimic as a task-aligned perturbation mechanism.
Here we formalize this design choice at the level of feasible perturbation classes.

Let $\Delta_{\mathrm{batt}}$ denote the set of perturbations that can be physically realized by the battery under capacity, power, state-of-charge, throughput, and cost constraints.
A generic battery-induced perturbation satisfies
\begin{equation}
\label{eq:app_generic_battery_perturbation}
    \delta_{\pi_\theta}
    \in
    \Delta_{\mathrm{batt}}.
\end{equation}
Signature mimic restricts this feasible set to perturbations generated by replaying trajectories from a real appliance-signature library $\mathcal{S}$:
\begin{equation}
\label{eq:app_signature_class}
    \Delta_{\mathrm{sig}}
    =
    \left\{
        \delta \in \Delta_{\mathrm{batt}}
        :
        \delta
        \text{ is induced by replaying a selected signature }
        s\in\mathcal{S}
    \right\}.
\end{equation}
Thus,
\begin{equation}
\label{eq:app_signature_subset}
    \Delta_{\mathrm{sig}}
    \subseteq
    \Delta_{\mathrm{batt}}.
\end{equation}

This restriction does not make the perturbation class more general.
Instead, it makes the perturbations more task-aligned.
The goal is not to maximize arbitrary distortion of the aggregate signal, but to introduce plausible appliance-like decoys that interfere with NILM disaggregation.
Compared with unstructured random perturbations sampled from $\Delta_{\mathrm{batt}}$, perturbations in $\Delta_{\mathrm{sig}}$ are constructed from real appliance trajectories and therefore retain appliance-like temporal structure.

This provides a formal interpretation of signature mimic as a structured adversarial perturbation class tailored to NILM inference.
The effectiveness of this design choice is evaluated empirically through comparisons with random or unstructured perturbation baselines and through ablations over the signature library.

\section{Algorithm}
\label{app:algorithm}
The full training procedure is summarized in Algorithm~\ref{alg:defender_train}. 

\begin{algorithm}[t]
\caption{Proxy-Guided HRL Training for Signature Mimicry}
\label{alg:defender_train}
\textbf{Input:} Fixed privacy probe \(f_\eta\), signature library \(\mathcal{S}\), battery environment \(\mathcal{E}\) \\
\textbf{Output:} Trained manager policy \(\pi_\theta\)

\begin{algorithmic}[1]
\STATE Initialize manager policy \(\pi_\theta\)
\FOR{each training episode}
    \STATE Reset environment and battery state
    \WHILE{episode not finished}
        \STATE Observe state \(s_t\)
        \STATE Sample high-level action \(a_t^{\mathrm{high}}\sim\pi_\theta(\cdot|s_t)\)
        \IF{\(a_t^{\mathrm{high}}=0\)}
            \STATE Apply no mimic action and advance one step
        \ELSE
            \STATE Select the signature and replay direction from \(a_t^{\mathrm{high}}\)
            \STATE Replay the selected signature using the deterministic executor subject to hard battery constraints
        \ENDIF
        \STATE Construct the resulting masked window \(\mathbf{x}^{\pi_\theta}_{t}\)
        \STATE Construct the corresponding unmodified window \(\mathbf{x}^{0}_{t}\)
        \STATE Compute privacy reward using Eq.~\eqref{eq:privacy_reward}
        \STATE Store transition in the rollout buffer
    \ENDWHILE
    \STATE Update manager policy \(\pi_\theta\) using PPO
\ENDFOR
\STATE Return trained policy \(\pi_\theta\)
\end{algorithmic}
\end{algorithm}

\section{Broader Impact}
\label{app:broader_impact}

This work develops a privacy defense for smart meter users against appliance-level inference 
attacks, aiming to restore user control over fine-grained behavioral information that smart 
meter deployments would otherwise expose. The intended societal benefit is improved residential 
energy privacy without requiring changes to the metering infrastructure or assumptions about 
the deployed inference model.

We acknowledge two potential concerns.
\emph{Misuse risk.} The same proxy-guided framework that learns to inject misleading appliance 
signatures could in principle be repurposed to obscure malicious load patterns, such as energy 
theft, from utility-side monitoring. We note that the framework operates under hard battery 
feasibility constraints and does not alter the user's actual energy consumption—only the 
reported aggregate trajectory. Distinguishing legitimate privacy protection from concealment 
of misuse requires utility-side detection mechanisms and is a policy-level question beyond the 
scope of this work.
\emph{Grid-level effects.} Wide-scale deployment of battery-based load shaping interacts with 
grid demand forecasting and could affect distribution-network operations if adopted at scale. 
Coordinated deployment protocols and forecasting-aware extensions are an important direction 
for practical adoption.

\section{Experiment Settings}
\subsection{Dataset Description}
\label{app:dataset}
We evaluate our method on the UK-DALE dataset~\cite{kelly2015uk}, a widely used public dataset for NILM research. It contains appliance-level and aggregate power readings from multiple UK households. We use data from House~1, downsampled to 1-minute resolution, following prior works~\cite{zhang2024proactive,zhang2024privacy}. We extract 10 days of user load data (from March 19 to March 29, 2013) as the training set, and reserve a separate day (March 18, 2013) as the test set.
During attacker training, the 10-day data is further split into a training and a cross-validation subset (80:20). The test day is only used to evaluate the performance of the learned defender policy.

To assess cross-dataset applicability, we additionally conduct experiments on the REDD dataset~\cite{kolter2011redd}. 
REDD contains aggregate and appliance-level power measurements from multiple residential buildings in the United States.
Similar to the UK-DALE setup, we use continuous data from Building~2, downsampled to 1-minute resolution. We select 11 consecutive days, using the first 10 days (from April~19 to April~29, 2011) for training and the 11th day (April~30, 2011) for testing. 
The REDD dataset is used to evaluate how well the proposed defense framework performs when facing a range of mainstream NILM attackers under a different data distribution.

\subsection{Appliance Signature Selection}
\label{app:signature}

We construct a trajectory-level appliance signature library $\mathcal{S}$ from the UK-DALE House~1 dataset, which contains power traces of 37 labeled household appliances. 
All candidate and selected signatures are extracted exclusively from the UK-DALE House 1 training period, with no temporal overlap with any evaluation day. Once constructed, the library is frozen and reused without target-specific adaptation.
The purpose of this library is to provide realistic appliance-level usage patterns that can be replayed by the battery controller during mimicry. 
Since NILM attackers often rely on distinctive appliance activation patterns in the aggregate load, we focus on intermittent high-power appliances whose signatures are visually and statistically salient, such as kettles, toasters, microwaves, ovens, washer--dryers, and dishwashers.

\paragraph{Candidate event extraction.}
We first identify the top-6 energy-consuming intermittent high-power appliances and extract candidate usage trajectories from a 10-day period. 
For each appliance, we apply a threshold-based event segmentation scheme. 
A device-specific activation threshold is used to determine whether the appliance is operating. 
The appliance-level power sequence is scanned sequentially: when the power first exceeds the threshold, a new event is initiated and its start time is recorded. 
Consecutive samples above the threshold are grouped into the same event, forming a continuous usage trajectory. 
The event is terminated once the power falls below the threshold. 
To remove noise and spurious activations, events shorter than a predefined minimum duration are discarded. 
This process yields 188 candidate appliance usage trajectories.

\paragraph{Filtering and ranking.}
To keep the mimic actions physically meaningful and to simplify the reinforcement learning action space, we further filter the candidate trajectories by duration and power magnitude. 
Specifically, we retain events whose duration is between 5 and 35 minutes and whose median power lies between 300 and 2500~W. 
These constraints remove extremely short, low-power, or unusually long events that are less suitable for battery-based replay.

The remaining candidates are ranked according to two criteria. 
First, we prioritize trajectories with larger standard deviation, as they exhibit stronger temporal variation and therefore provide more distinctive usage patterns. 
Second, we rank by median power, since higher-power events are more visible in the aggregate load and are more likely to affect NILM inference. 
To avoid the final library being dominated by a single appliance type, we impose a balancing constraint and retain at most two signatures per appliance category. 
Signatures are then selected greedily from the ordered candidate list until the final library contains 10 representative trajectories.

\paragraph{Final signature library.}
The resulting library $\mathcal{S}$ contains 10 appliance signatures covering short high-power bursts and medium-duration usage patterns. 
Typical examples include kettle, toaster, microwave, washer-dryer, and dishwasher signatures. 
All selected signatures have durations between 5 and 35 minutes, with most concentrated in the 5--10 minute range. 
In terms of power magnitude, most signatures fall between 1500 and 2500~W, corresponding to common high-power intermittent household appliances.

\begin{figure}
    \centering
    \includegraphics[width=0.99\linewidth]{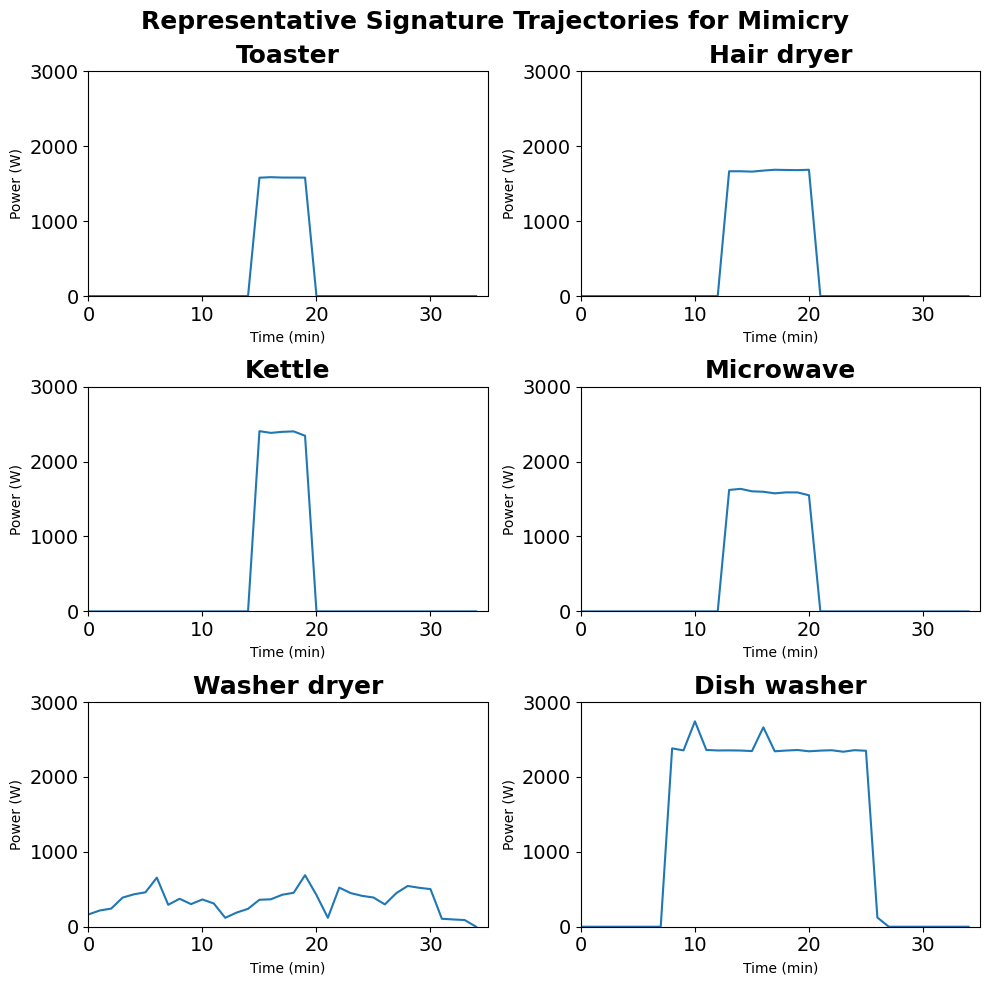}
    \caption{Representative signature trajectories in the constructed signature library. All trajectories are padded to a unified length of 35 minutes for visualization. The signatures exhibit diverse temporal lengths, power magnitudes, and usage patterns, illustrating the diversity and realism of the signature library used for mimicry.}
    \label{fig:signature}
\end{figure}

Figure~\ref{fig:signature} visualizes representative trajectories from the final mimic library. 
The selected signatures exhibit diverse temporal lengths, power magnitudes, and usage patterns, supporting realistic appliance-level mimicry under battery control.

\subsection{Privacy Probe Training}
\label{app:privacy_probe}
The architecture of the Transformer-based Seq2Seq privacy probe model consists of a linear embedding layer projecting input to a 64-dimensional latent space, followed by sinusoidal positional encoding. The encoder comprises 3 Transformer layers with 4 attention heads and a dropout rate of 0.1. A final linear projection maps each hidden state to a 1D power prediction. 

The LSTM probe uses a two-layer LSTM encoder with a hidden size of 64. A single-channel input sequence is first processed by the stacked recurrent layers, which output a 64-dimensional hidden representation at each time step. A final linear projection maps each hidden state to a scalar power estimate. 

The 1D-CNN probe is built with three temporal convolutional layers. After reshaping the input to channel-first format, the model applies two hidden Conv1D blocks (64 channels, kernel size 3), each followed by ReLU activation, and a final Conv1D layer that maps features to a single output channel. Symmetric padding is used to preserve sequence length throughout the network, and the output is reshaped back to time-first format. 


We first generate candidate 60-minute segments via repeated sliding-window passes (stride = 30 minutes) over the 14,400-minute (10-day) period, with each pass beginning at a randomly sampled starting offset uniformly drawn from [0, 1439] and proceeding to the end of the sequence. Within each pass, the first 80\% of generated segments (by temporal order) are tentatively tagged as the training set and the last 20\% as the validation set. To eliminate leakage arising from the randomized offsets, we discard any training-tagged segment whose end time extends into the final two validation days(i.e.,  exceeds 11,520 minutes), and any validation-tagged segment whose start time extends into the first eight training days (i.e., precedes 11,520 minutes). This guarantees strict non-overlap in absolute time between the training and validation sets. The procedure repeats across independently sampled offsets until 5,000 segments are collected for each set.
The remaining one-day data is reserved exclusively for evaluation. The model is trained using mean squared error (MSE) loss as loss function.

\subsection{Defender Training Parameters and SM environment Settings}
\label{app:defender_setting}
The PPO defender training proceeds for 300 PPO episodes, with 1,440 time steps per episode (corresponding to one full day). The PPO optimizer uses a learning rate of
$2 \times 10^{-5}$, a discount factor $\gamma = 0.99$, GAE parameter $\lambda = 0.95$. We adopt a rollout horizon of 4,096 steps, 4 optimization epochs per update, a minibatch size of 2,048, and a clipping range of 0.1. Entropy regularization is enabled with a coefficient of 0.01 to encourage early exploration, and a target KL divergence of 0.01 is used to stabilize policy updates.

The defender operates within a simulated battery and SM environment with the following constraints: a maximum capacity of 8 kWh, peak charge/discharge power of 4 kW, and a SoC constraint between 10\% and 90\%. The SoC is initialized to 50\% at the beginning of each episode. All actions are applied at 1-minute resolution.

\subsection{Computational resources.}
All experiments were conducted on a single NVIDIA RTX 3050 Ti GPU. Defender training 
takes approximately 1.2 hours per run; full reproduction of the main results 
(both datasets, five seeds, six attackers) requires roughly 30 GPU-hours.
\section{Original Experiment results}
\subsection{Detailed Main Results}
\label{app:detailed_main_results}
\begin{table*}[t]
\centering
\caption{Detailed Raw, Masked, Delta, and p-value results on UK-DALE over five random seeds.}
\label{tab:attacker_ukdale}
\resizebox{\textwidth}{!}{
\begingroup
\fontsize{12pt}{14pt}\selectfont
\renewcommand{\arraystretch}{1.85}
\newcommand{\num}[1]{{\fontsize{13.5pt}{14pt}\selectfont #1}}
\newcommand{\ms}[2]{\num{#1$\pm$#2}}

\begin{tabular}{ll cccc cccc cccc cccc}
\toprule

\multirow{2}{*}{\rotatebox{90}{\textbf{Attacker}}} 
& \multirow{2}{*}{\textbf{Appliance}} 
  & \multicolumn{4}{c}{Raw Data} 
    & \multicolumn{4}{c}{Masked Data} 
    & \multicolumn{4}{c}{\textbf{Delta (Masked - Raw)}} 
    & \multicolumn{4}{c}{p-value} \\

\cmidrule(lr){3-6}
\cmidrule(lr){7-10}
\cmidrule(lr){11-14}
\cmidrule(lr){15-18}

&
& RMSE & MAE & SAE & F1 
& RMSE & MAE & SAE & F1 
& RMSE & MAE & SAE & F1 
& RMSE & MAE & SAE & F1 \\
\hline

\multirow{5}{*}{\rotatebox{90}{CO}} 
& Aggregate
& \num{176.97} & \num{31.53} & \num{0.07} & --
& \ms{393.16}{26.34} & \ms{194.98}{7.13} & \ms{0.84}{0.07} & --
& \ms{216.18}{26.34} & \ms{163.44}{7.13} & \ms{0.77}{0.07} & --
& \num{0.00} & \num{0.00} & \num{0.00} & -- \\

& Fridge freezer
& \num{79.24} & \num{41.01} & \num{0.81} & \num{0.70}
& \ms{82.41}{1.33} & \ms{40.33}{0.86} & \ms{0.84}{0.01} & \ms{0.16}{0.03}
& \ms{3.17}{1.33} & \ms{-0.68}{0.86} & \ms{0.02}{0.01} & \ms{-0.54}{0.03}
& \num{0.01} & \num{0.19} & \num{0.02} & \num{0.00} \\

& Kettle
& \num{86.62} & \num{4.85} & \num{0.05} & \num{0.95}
& \ms{274.23}{11.35} & \ms{33.14}{2.83} & \ms{0.59}{0.29} & \ms{0.25}{0.14}
& \ms{187.61}{11.35} & \ms{28.29}{2.83} & \ms{0.54}{0.29} & \ms{-0.70}{0.14}
& \num{0.00} & \num{0.00} & \num{0.02} & \num{0.00} \\

& Microwave
& \num{41.76} & \num{2.37} & \num{0.02} & \num{0.96}
& \ms{167.45}{24.36} & \ms{18.83}{5.02} & \ms{0.49}{0.29} & \ms{0.08}{0.11}
& \ms{125.70}{24.36} & \ms{16.46}{5.02} & \ms{0.47}{0.29} & \ms{-0.87}{0.11}
& \num{0.00} & \num{0.00} & \num{0.03} & \num{0.00} \\

& Toaster
& \num{94.89} & \num{6.40} & \num{0.21} & \num{0.89}
& \ms{188.11}{9.56} & \ms{20.57}{2.22} & \ms{0.90}{0.12} & \ms{0.00}{0.00}
& \ms{93.23}{9.56} & \ms{14.17}{2.22} & \ms{0.69}{0.12} & \ms{-0.89}{0.00}
& \num{0.00} & \num{0.00} & \num{0.00} & \num{0.00} \\

\hline

\multirow{5}{*}{\rotatebox{90}{FHMM}} 
& Aggregate
& \num{150.54} & \num{20.62} & \num{0.02} & --
& \ms{389.00}{26.01} & \ms{183.97}{7.81} & \ms{0.74}{0.07} & --
& \ms{238.46}{26.01} & \ms{163.35}{7.81} & \ms{0.72}{0.07} & --
& \num{0.00} & \num{0.00} & \num{0.00} & -- \\

& Fridge freezer
& \num{77.11} & \num{39.07} & \num{0.78} & \num{0.70}
& \ms{82.87}{2.22} & \ms{40.64}{1.26} & \ms{0.83}{0.02} & \ms{0.16}{0.03}
& \ms{5.76}{2.22} & \ms{1.57}{1.26} & \ms{0.05}{0.02} & \ms{-0.54}{0.03}
& \num{0.01} & \num{0.07} & \num{0.01} & \num{0.00} \\

& Kettle
& \num{86.62} & \num{4.85} & \num{0.05} & \num{0.95}
& \ms{274.23}{11.35} & \ms{33.14}{2.83} & \ms{0.59}{0.29} & \ms{0.25}{0.14}
& \ms{187.61}{11.35} & \ms{28.29}{2.83} & \ms{0.54}{0.29} & \ms{-0.70}{0.14}
& \num{0.00} & \num{0.00} & \num{0.02} & \num{0.00} \\

& Microwave
& \num{41.76} & \num{2.37} & \num{0.02} & \num{0.96}
& \ms{167.45}{24.36} & \ms{18.83}{5.02} & \ms{0.49}{0.29} & \ms{0.08}{0.11}
& \ms{125.70}{24.36} & \ms{16.46}{5.02} & \ms{0.47}{0.29} & \ms{-0.87}{0.11}
& \num{0.00} & \num{0.00} & \num{0.03} & \num{0.00} \\

& Toaster
& \num{94.89} & \num{6.40} & \num{0.21} & \num{0.89}
& \ms{188.11}{9.56} & \ms{20.57}{2.22} & \ms{0.90}{0.12} & \ms{0.00}{0.00}
& \ms{93.23}{9.56} & \ms{14.17}{2.22} & \ms{0.69}{0.12} & \ms{-0.89}{0.00}
& \num{0.00} & \num{0.00} & \num{0.00} & \num{0.00} \\

\hline

\multirow{5}{*}{\rotatebox{90}{DAE}} 
& Aggregate
& \num{46.09} & \num{22.38} & \num{0.01} & --
& \ms{471.59}{36.74} & \ms{224.61}{11.86} & \ms{0.46}{0.08} & --
& \ms{425.50}{36.74} & \ms{202.23}{11.86} & \ms{0.44}{0.08} & --
& \num{0.00} & \num{0.00} & \num{0.00} & -- \\

& Fridge freezer
& \num{62.07} & \num{28.91} & \num{0.10} & \num{0.74}
& \ms{78.79}{1.08} & \ms{40.09}{1.17} & \ms{0.63}{0.03} & \ms{0.24}{0.02}
& \ms{16.72}{1.08} & \ms{11.18}{1.17} & \ms{0.53}{0.03} & \ms{-0.50}{0.02}
& \num{0.00} & \num{0.00} & \num{0.00} & \num{0.00} \\

& Kettle
& \num{172.13} & \num{35.48} & \num{0.25} & \num{0.72}
& \ms{323.42}{16.39} & \ms{70.15}{8.34} & \ms{0.57}{0.31} & \ms{0.24}{0.09}
& \ms{151.29}{16.39} & \ms{34.67}{8.34} & \ms{0.32}{0.31} & \ms{-0.48}{0.09}
& \num{0.00} & \num{0.00} & \num{0.11} & \num{0.00} \\

& Microwave
& \num{121.10} & \num{33.13} & \num{1.29} & \num{0.31}
& \ms{198.14}{15.75} & \ms{50.26}{5.29} & \ms{2.22}{0.21} & \ms{0.16}{0.05}
& \ms{77.04}{15.75} & \ms{17.13}{5.29} & \ms{0.93}{0.21} & \ms{-0.15}{0.05}
& \num{0.00} & \num{0.00} & \num{0.00} & \num{0.00} \\

& Toaster
& \num{151.07} & \num{36.44} & \num{0.74} & \num{0.57}
& \ms{216.61}{17.21} & \ms{49.77}{5.81} & \ms{0.67}{0.44} & \ms{0.15}{0.13}
& \ms{65.54}{17.21} & \ms{13.33}{5.81} & \ms{-0.06}{0.44} & \ms{-0.42}{0.13}
& \num{0.00} & \num{0.01} & \num{0.79} & \num{0.00} \\

\hline

\multirow{5}{*}{\rotatebox{90}{S2P}} 
& Aggregate
& \num{26.62} & \num{12.72} & \num{0.00} & --
& \ms{489.24}{34.84} & \ms{233.44}{12.27} & \ms{0.47}{0.08} & --
& \ms{462.62}{34.84} & \ms{220.71}{12.27} & \ms{0.46}{0.08} & --
& \num{0.00} & \num{0.00} & \num{0.00} & -- \\

& Fridge freezer
& \num{38.28} & \num{19.45} & \num{0.10} & \num{0.87}
& \ms{80.18}{2.13} & \ms{40.86}{0.62} & \ms{0.67}{0.02} & \ms{0.17}{0.03}
& \ms{41.90}{2.13} & \ms{21.41}{0.62} & \ms{0.57}{0.02} & \ms{-0.70}{0.03}
& \num{0.00} & \num{0.00} & \num{0.00} & \num{0.00} \\

& Kettle
& \num{127.35} & \num{15.49} & \num{0.36} & \num{0.91}
& \ms{278.11}{18.24} & \ms{44.24}{4.88} & \ms{0.28}{0.18} & \ms{0.21}{0.17}
& \ms{150.75}{18.24} & \ms{28.75}{4.88} & \ms{-0.08}{0.18} & \ms{-0.70}{0.17}
& \num{0.00} & \num{0.00} & \num{0.42} & \num{0.00} \\

& Microwave
& \num{150.39} & \num{24.99} & \num{1.14} & \num{0.37}
& \ms{223.62}{21.83} & \ms{43.87}{7.30} & \ms{2.12}{0.54} & \ms{0.18}{0.05}
& \ms{73.23}{21.83} & \ms{18.88}{7.30} & \ms{0.97}{0.54} & \ms{-0.19}{0.05}
& \num{0.00} & \num{0.01} & \num{0.02} & \num{0.00} \\

& Toaster
& \num{80.27} & \num{10.80} & \num{0.03} & \num{0.94}
& \ms{224.70}{25.56} & \ms{41.86}{8.95} & \ms{0.67}{0.53} & \ms{0.11}{0.07}
& \ms{144.44}{25.56} & \ms{31.06}{8.95} & \ms{0.64}{0.53} & \ms{-0.83}{0.07}
& \num{0.00} & \num{0.00} & \num{0.07} & \num{0.00} \\

\hline

\multirow{5}{*}{\rotatebox{90}{ELECTRICITY}} 
& Aggregate
& \num{16.20} & \num{14.11} & \num{0.05} & --
& \ms{495.12}{35.71} & \ms{250.55}{12.46} & \ms{0.53}{0.08} & --
& \ms{478.92}{35.71} & \ms{236.43}{12.46} & \ms{0.48}{0.08} & --
& \num{0.00} & \num{0.00} & \num{0.00} & -- \\

& Fridge freezer
& \num{33.13} & \num{17.19} & \num{0.12} & \num{0.84}
& \ms{83.88}{1.65} & \ms{42.80}{0.63} & \ms{1.00}{0.01} & \ms{0.07}{0.03}
& \ms{50.75}{1.65} & \ms{25.61}{0.63} & \ms{0.88}{0.01} & \ms{-0.77}{0.03}
& \num{0.00} & \num{0.00} & \num{0.00} & \num{0.00} \\

& Kettle
& \num{88.04} & \num{8.37} & \num{0.21} & \num{0.94}
& \ms{323.55}{44.29} & \ms{53.82}{11.85} & \ms{0.24}{0.14} & \ms{0.24}{0.09}
& \ms{235.51}{44.29} & \ms{45.45}{11.85} & \ms{0.03}{0.14} & \ms{-0.70}{0.09}
& \num{0.00} & \num{0.00} & \num{0.68} & \num{0.00} \\

& Microwave
& \num{172.84} & \num{25.86} & \num{1.20} & \num{0.50}
& \ms{188.11}{25.85} & \ms{35.08}{8.12} & \ms{0.96}{0.69} & \ms{0.19}{0.06}
& \ms{15.27}{25.85} & \ms{9.22}{8.12} & \ms{-0.24}{0.69} & \ms{-0.31}{0.06}
& \num{0.30} & \num{0.09} & \num{0.53} & \num{0.00} \\

& Toaster
& \num{129.56} & \num{21.01} & \num{0.14} & \num{0.76}
& \ms{249.34}{28.75} & \ms{51.65}{10.71} & \ms{0.84}{0.53} & \ms{0.20}{0.11}
& \ms{119.79}{28.75} & \ms{30.64}{10.71} & \ms{0.70}{0.53} & \ms{-0.56}{0.11}
& \num{0.00} & \num{0.00} & \num{0.06} & \num{0.00} \\

\hline

\multirow{5}{*}{\rotatebox{90}{BERT4NILM}} 
& Aggregate
& \num{11.86} & \num{11.32} & \num{0.04} & --
& \ms{491.00}{34.54} & \ms{247.27}{12.20} & \ms{0.53}{0.08} & --
& \ms{479.14}{34.54} & \ms{235.95}{12.20} & \ms{0.48}{0.08} & --
& \num{0.00} & \num{0.00} & \num{0.00} & -- \\

& Fridge freezer
& \num{31.46} & \num{16.22} & \num{0.09} & \num{0.84}
& \ms{83.43}{0.46} & \ms{42.73}{0.41} & \ms{1.02}{0.01} & \ms{0.05}{0.02}
& \ms{51.97}{0.46} & \ms{26.50}{0.41} & \ms{0.92}{0.01} & \ms{-0.79}{0.02}
& \num{0.00} & \num{0.00} & \num{0.00} & \num{0.00} \\

& Kettle
& \num{55.75} & \num{6.79} & \num{0.17} & \num{0.97}
& \ms{336.92}{43.63} & \ms{57.87}{12.34} & \ms{0.31}{0.07} & \ms{0.25}{0.12}
& \ms{281.17}{43.63} & \ms{51.08}{12.34} & \ms{0.14}{0.07} & \ms{-0.72}{0.12}
& \num{0.00} & \num{0.00} & \num{0.02} & \num{0.00} \\

& Microwave
& \num{164.56} & \num{26.23} & \num{0.94} & \num{0.52}
& \ms{193.75}{22.96} & \ms{44.02}{8.14} & \ms{0.67}{0.41} & \ms{0.18}{0.04}
& \ms{29.19}{22.96} & \ms{17.79}{8.14} & \ms{-0.26}{0.41} & \ms{-0.35}{0.04}
& \num{0.06} & \num{0.01} & \num{0.27} & \num{0.00} \\

& Toaster
& \num{121.86} & \num{16.81} & \num{0.34} & \num{0.76}
& \ms{262.04}{32.90} & \ms{51.07}{11.57} & \ms{0.94}{0.84} & \ms{0.20}{0.11}
& \ms{140.18}{32.90} & \ms{34.26}{11.57} & \ms{0.60}{0.84} & \ms{-0.57}{0.11}
& \num{0.00} & \num{0.00} & \num{0.23} & \num{0.00} \\

\hline

\end{tabular}
\endgroup
}
\end{table*}

\begin{table*}[t]
\centering
\caption{Detailed Raw, Masked, Delta, and p-value results on REDD over five random seeds.}
\label{tab:attacker_redd}
\resizebox{\textwidth}{!}{
\begingroup
\fontsize{12pt}{14pt}\selectfont
\renewcommand{\arraystretch}{1.85}
\newcommand{\num}[1]{{\fontsize{13.5pt}{14pt}\selectfont #1}}
\newcommand{\ms}[2]{\num{#1$\pm$#2}}

\begin{tabular}{ll cccc cccc cccc cccc}
\toprule

\multirow{2}{*}{\rotatebox{90}{\textbf{Attacker}}} 
& \multirow{2}{*}{\textbf{Appliance}} 
  & \multicolumn{4}{c}{Raw Data} 
    & \multicolumn{4}{c}{Masked Data} 
    & \multicolumn{4}{c}{\textbf{Delta (Masked - Raw)}} 
    & \multicolumn{4}{c}{p-value} \\

\cmidrule(lr){3-6}
\cmidrule(lr){7-10}
\cmidrule(lr){11-14}
\cmidrule(lr){15-18}

&
& RMSE & MAE & SAE & F1 
& RMSE & MAE & SAE & F1 
& RMSE & MAE & SAE & F1 
& RMSE & MAE & SAE & F1 \\
\hline

\multirow{6}{*}{\rotatebox{90}{CO}} 
& Aggregate
& \num{69.69} & \num{13.72} & \num{0.03} & --
& \ms{444.12}{44.37} & \ms{281.58}{13.68} & \ms{0.56}{0.09} & --
& \ms{374.44}{44.37} & \ms{267.86}{13.68} & \ms{0.53}{0.09} & --
& \num{0.00} & \num{0.00} & \num{0.00} & -- \\

& Fridge
& \num{77.27} & \num{56.16} & \num{0.33} & \num{0.87}
& \ms{118.83}{1.01} & \ms{83.55}{0.83} & \ms{0.92}{0.02} & \ms{0.19}{0.02}
& \ms{41.56}{1.01} & \ms{27.39}{0.83} & \ms{0.59}{0.02} & \ms{-0.68}{0.02}
& \num{0.00} & \num{0.00} & \num{0.00} & \num{0.00} \\

& Light
& \num{48.79} & \num{22.11} & \num{0.26} & \num{0.22}
& \ms{46.99}{1.33} & \ms{22.19}{0.88} & \ms{0.73}{0.04} & \ms{0.06}{0.07}
& \ms{-1.80}{1.33} & \ms{0.08}{0.88} & \ms{0.48}{0.04} & \ms{-0.16}{0.07}
& \num{0.04} & \num{0.84} & \num{0.00} & \num{0.01} \\

& Microwave
& \num{138.65} & \num{33.33} & \num{2.89} & \num{0.06}
& \ms{326.20}{21.71} & \ms{81.38}{9.53} & \ms{8.62}{1.24} & \ms{0.00}{0.01}
& \ms{187.55}{21.71} & \ms{48.05}{9.53} & \ms{5.73}{1.24} & \ms{-0.05}{0.01}
& \num{0.00} & \num{0.00} & \num{0.00} & \num{0.00} \\

& Sockets 1
& \num{33.86} & \num{8.92} & \num{0.44} & \num{0.20}
& \ms{33.29}{0.13} & \ms{4.71}{0.42} & \ms{0.88}{0.04} & \ms{0.13}{0.18}
& \ms{-0.57}{0.13} & \ms{-4.21}{0.42} & \ms{0.44}{0.04} & \ms{-0.06}{0.18}
& \num{0.00} & \num{0.00} & \num{0.00} & \num{0.49} \\

& Sockets 2
& \num{148.77} & \num{98.30} & \num{7.95} & \num{0.07}
& \ms{75.18}{3.35} & \ms{13.35}{2.69} & \ms{0.68}{0.26} & \ms{0.06}{0.08}
& \ms{-73.59}{3.35} & \ms{-84.95}{2.69} & \ms{-7.27}{0.26} & \ms{-0.01}{0.08}
& \num{0.00} & \num{0.00} & \num{0.00} & \num{0.85} \\

\hline

\multirow{6}{*}{\rotatebox{90}{FHMM}} 
& Aggregate
& \num{73.52} & \num{41.97} & \num{0.08} & --
& \ms{446.71}{81.86} & \ms{262.56}{25.94} & \ms{0.34}{0.13} & --
& \ms{373.19}{81.86} & \ms{220.59}{25.94} & \ms{0.26}{0.13} & --
& \num{0.00} & \num{0.00} & \num{0.01} & -- \\

& Fridge
& \num{66.89} & \num{40.61} & \num{0.32} & \num{0.76}
& \ms{102.32}{1.56} & \ms{79.73}{1.13} & \ms{0.45}{0.02} & \ms{0.54}{0.01}
& \ms{35.43}{1.56} & \ms{39.12}{1.13} & \ms{0.13}{0.02} & \ms{-0.21}{0.01}
& \num{0.00} & \num{0.00} & \num{0.00} & \num{0.00} \\

& Light
& \num{96.23} & \num{71.92} & \num{2.84} & \num{0.30}
& \ms{45.07}{2.60} & \ms{20.90}{1.71} & \ms{0.81}{0.11} & \ms{0.07}{0.13}
& \ms{-51.16}{2.60} & \ms{-51.02}{1.71} & \ms{-2.04}{0.11} & \ms{-0.23}{0.13}
& \num{0.00} & \num{0.00} & \num{0.00} & \num{0.02} \\

& Microwave
& \num{141.72} & \num{34.66} & \num{3.06} & \num{0.06}
& \ms{349.26}{51.61} & \ms{89.13}{18.17} & \ms{9.62}{2.37} & \ms{0.00}{0.01}
& \ms{207.53}{51.61} & \ms{54.47}{18.17} & \ms{6.56}{2.37} & \ms{-0.05}{0.01}
& \num{0.00} & \num{0.00} & \num{0.00} & \num{0.00} \\

& Sockets 1
& \num{33.50} & \num{5.43} & \num{0.81} & \num{0.01}
& \ms{33.38}{0.06} & \ms{4.91}{0.32} & \ms{0.91}{0.05} & \ms{0.03}{0.07}
& \ms{-0.12}{0.06} & \ms{-0.52}{0.32} & \ms{0.10}{0.05} & \ms{0.03}{0.07}
& \num{0.01} & \num{0.02} & \num{0.01} & \num{0.43} \\

& Sockets 2
& \num{80.39} & \num{17.65} & \num{0.61} & \num{0.40}
& \ms{73.40}{1.10} & \ms{11.34}{0.50} & \ms{0.70}{0.03} & \ms{0.05}{0.06}
& \ms{-6.99}{1.10} & \ms{-6.31}{0.50} & \ms{0.09}{0.03} & \ms{-0.35}{0.06}
& \num{0.00} & \num{0.00} & \num{0.00} & \num{0.00} \\

\hline

\multirow{6}{*}{\rotatebox{90}{DAE}} 
& Aggregate
& \num{28.92} & \num{15.25} & \num{0.01} & --
& \ms{480.67}{62.38} & \ms{285.13}{16.30} & \ms{0.37}{0.11} & --
& \ms{451.75}{62.38} & \ms{269.88}{16.30} & \ms{0.35}{0.11} & --
& \num{0.00} & \num{0.00} & \num{0.00} & -- \\

& Fridge
& \num{29.06} & \num{14.88} & \num{0.01} & \num{0.95}
& \ms{114.22}{3.72} & \ms{82.69}{2.77} & \ms{0.78}{0.04} & \ms{0.26}{0.06}
& \ms{85.16}{3.72} & \ms{67.81}{2.77} & \ms{0.77}{0.04} & \ms{-0.69}{0.06}
& \num{0.00} & \num{0.00} & \num{0.00} & \num{0.00} \\

& Light
& \num{33.81} & \num{15.63} & \num{0.03} & \num{0.30}
& \ms{45.57}{12.39} & \ms{14.67}{3.43} & \ms{0.47}{0.19} & \ms{0.10}{0.15}
& \ms{11.76}{12.39} & \ms{-0.96}{3.43} & \ms{0.44}{0.19} & \ms{-0.19}{0.15}
& \num{0.10} & \num{0.57} & \num{0.01} & \num{0.04} \\

& Microwave
& \num{26.07} & \num{6.01} & \num{0.52} & \num{0.75}
& \ms{118.84}{23.04} & \ms{26.41}{7.32} & \ms{2.50}{0.92} & \ms{0.00}{0.00}
& \ms{92.77}{23.04} & \ms{20.40}{7.32} & \ms{1.99}{0.92} & \ms{-0.75}{0.00}
& \num{0.00} & \num{0.00} & \num{0.01} & -- \\

& Sockets 1
& \num{33.06} & \num{8.01} & \num{0.53} & \num{0.24}
& \ms{34.32}{0.11} & \ms{13.06}{0.46} & \ms{1.98}{0.09} & \ms{0.30}{0.06}
& \ms{1.26}{0.11} & \ms{5.05}{0.46} & \ms{1.45}{0.09} & \ms{0.05}{0.06}
& \num{0.00} & \num{0.00} & \num{0.00} & \num{0.10} \\

& Sockets 2
& \num{37.49} & \num{12.42} & \num{0.21} & \num{0.28}
& \ms{226.49}{53.25} & \ms{73.36}{10.66} & \ms{4.30}{1.05} & \ms{0.09}{0.09}
& \ms{189.00}{53.25} & \ms{60.95}{10.66} & \ms{4.09}{1.05} & \ms{-0.20}{0.09}
& \num{0.00} & \num{0.00} & \num{0.00} & \num{0.01} \\

\hline

\multirow{6}{*}{\rotatebox{90}{S2P}} 
& Aggregate
& \num{39.09} & \num{12.63} & \num{0.01} & --
& \ms{468.86}{32.56} & \ms{274.56}{14.17} & \ms{0.44}{0.09} & --
& \ms{429.78}{32.56} & \ms{261.93}{14.17} & \ms{0.43}{0.09} & --
& \num{0.00} & \num{0.00} & \num{0.00} & -- \\

& Fridge
& \num{34.43} & \num{16.12} & \num{0.08} & \num{0.95}
& \ms{117.25}{2.33} & \ms{84.07}{2.18} & \ms{0.87}{0.03} & \ms{0.15}{0.05}
& \ms{82.82}{2.33} & \ms{67.95}{2.18} & \ms{0.79}{0.03} & \ms{-0.81}{0.05}
& \num{0.00} & \num{0.00} & \num{0.00} & \num{0.00} \\

& Light
& \num{17.71} & \num{7.92} & \num{0.06} & \num{0.46}
& \ms{40.57}{0.84} & \ms{18.38}{0.52} & \ms{0.80}{0.11} & \ms{0.14}{0.17}
& \ms{22.86}{0.84} & \ms{10.46}{0.52} & \ms{0.74}{0.11} & \ms{-0.33}{0.17}
& \num{0.00} & \num{0.00} & \num{0.00} & \num{0.01} \\

& Microwave
& \num{21.71} & \num{5.99} & \num{0.54} & \num{0.67}
& \ms{88.40}{14.50} & \ms{14.01}{2.84} & \ms{0.30}{0.24} & \ms{0.00}{0.00}
& \ms{66.69}{14.50} & \ms{8.02}{2.84} & \ms{-0.24}{0.24} & \ms{-0.67}{0.00}
& \num{0.00} & \num{0.00} & \num{0.09} & -- \\

& Sockets 1
& \num{35.96} & \num{9.22} & \num{0.65} & \num{0.26}
& \ms{40.08}{8.52} & \ms{7.18}{3.00} & \ms{0.58}{0.16} & \ms{0.06}{0.09}
& \ms{4.12}{8.52} & \ms{-2.04}{3.00} & \ms{-0.07}{0.16} & \ms{-0.20}{0.09}
& \num{0.34} & \num{0.20} & \num{0.38} & \num{0.01} \\

& Sockets 2
& \num{22.87} & \num{4.56} & \num{0.21} & \num{0.78}
& \ms{180.42}{60.98} & \ms{43.52}{16.65} & \ms{2.60}{1.66} & \ms{0.08}{0.09}
& \ms{157.54}{60.98} & \ms{38.96}{16.65} & \ms{2.39}{1.66} & \ms{-0.70}{0.09}
& \num{0.00} & \num{0.01} & \num{0.03} & \num{0.00} \\

\hline

\multirow{6}{*}{\rotatebox{90}{ELECTRICITY}} 
& Aggregate
& \num{7.72} & \num{6.00} & \num{0.02} & --
& \ms{510.80}{50.87} & \ms{301.77}{14.85} & \ms{0.40}{0.10} & --
& \ms{503.08}{50.87} & \ms{295.77}{14.85} & \ms{0.38}{0.10} & --
& \num{0.00} & \num{0.00} & \num{0.00} & -- \\

& Fridge
& \num{22.57} & \num{8.84} & \num{0.01} & \num{0.97}
& \ms{121.09}{1.54} & \ms{85.05}{2.07} & \ms{0.96}{0.03} & \ms{0.03}{0.06}
& \ms{98.52}{1.54} & \ms{76.21}{2.07} & \ms{0.95}{0.03} & \ms{-0.94}{0.06}
& \num{0.00} & \num{0.00} & \num{0.00} & \num{0.00} \\

& Light
& \num{19.34} & \num{7.93} & \num{0.08} & \num{0.40}
& \ms{38.21}{0.17} & \ms{11.57}{0.07} & \ms{0.60}{0.01} & \ms{0.01}{0.02}
& \ms{18.86}{0.17} & \ms{3.64}{0.07} & \ms{0.52}{0.01} & \ms{-0.40}{0.02}
& \num{0.00} & \num{0.00} & \num{0.00} & \num{0.00} \\

& Microwave
& \num{11.81} & \num{3.91} & \num{0.14} & \num{1.00}
& \ms{333.52}{55.63} & \ms{89.82}{22.03} & \ms{8.74}{3.02} & \ms{0.01}{0.01}
& \ms{321.71}{55.63} & \ms{85.92}{22.03} & \ms{8.60}{3.02} & \ms{-0.99}{0.01}
& \num{0.00} & \num{0.00} & \num{0.00} & \num{0.00} \\

& Sockets 1
& \num{7.60} & \num{3.85} & \num{0.05} & \num{0.04}
& \ms{37.14}{0.80} & \ms{18.88}{1.98} & \ms{2.70}{0.26} & \ms{0.27}{0.20}
& \ms{29.55}{0.80} & \ms{15.03}{1.98} & \ms{2.64}{0.26} & \ms{0.23}{0.20}
& \num{0.00} & \num{0.00} & \num{0.00} & \num{0.07} \\

& Sockets 2
& \num{12.32} & \num{5.17} & \num{0.17} & \num{0.71}
& \ms{92.20}{15.54} & \ms{19.80}{3.62} & \ms{0.44}{0.28} & \ms{0.05}{0.05}
& \ms{79.88}{15.54} & \ms{14.63}{3.62} & \ms{0.27}{0.28} & \ms{-0.66}{0.05}
& \num{0.00} & \num{0.00} & \num{0.10} & \num{0.00} \\

\hline

\multirow{6}{*}{\rotatebox{90}{BERT4NILM}} 
& Aggregate
& \num{4.22} & \num{3.24} & \num{0.01} & --
& \ms{498.50}{41.98} & \ms{300.22}{12.80} & \ms{0.43}{0.09} & --
& \ms{494.29}{41.98} & \ms{296.98}{12.80} & \ms{0.42}{0.09} & --
& \num{0.00} & \num{0.00} & \num{0.00} & -- \\

& Fridge
& \num{19.68} & \num{8.34} & \num{0.01} & \num{0.98}
& \ms{120.19}{1.14} & \ms{85.62}{1.58} & \ms{0.95}{0.03} & \ms{0.19}{0.03}
& \ms{100.51}{1.14} & \ms{77.29}{1.58} & \ms{0.94}{0.03} & \ms{-0.79}{0.03}
& \num{0.00} & \num{0.00} & \num{0.00} & \num{0.00} \\

& Light
& \num{19.84} & \num{7.96} & \num{0.03} & \num{0.43}
& \ms{38.14}{0.02} & \ms{11.36}{0.03} & \ms{0.60}{0.00} & \ms{0.01}{0.02}
& \ms{18.30}{0.02} & \ms{3.41}{0.03} & \ms{0.57}{0.00} & \ms{-0.42}{0.02}
& \num{0.00} & \num{0.00} & \num{0.00} & \num{0.00} \\

& Microwave
& \num{11.82} & \num{3.24} & \num{0.12} & \num{0.89}
& \ms{418.96}{29.23} & \ms{116.65}{13.85} & \ms{12.38}{1.89} & \ms{0.00}{0.01}
& \ms{407.14}{29.23} & \ms{113.41}{13.85} & \ms{12.25}{1.89} & \ms{-0.88}{0.01}
& \num{0.00} & \num{0.00} & \num{0.00} & \num{0.00} \\

& Sockets 1
& \num{11.10} & \num{3.99} & \num{0.19} & \num{0.04}
& \ms{34.00}{0.23} & \ms{8.08}{0.41} & \ms{1.67}{0.08} & \ms{0.00}{0.00}
& \ms{22.90}{0.23} & \ms{4.09}{0.41} & \ms{1.48}{0.08} & \ms{-0.04}{0.00}
& \num{0.00} & \num{0.00} & \num{0.00} & -- \\

& Sockets 2
& \num{28.64} & \num{9.27} & \num{0.62} & \num{0.52}
& \ms{74.28}{0.06} & \ms{20.34}{0.07} & \ms{1.91}{0.01} & \ms{0.00}{0.00}
& \ms{45.64}{0.06} & \ms{11.08}{0.07} & \ms{1.29}{0.01} & \ms{-0.52}{0.00}
& \num{0.00} & \num{0.00} & \num{0.00} & -- \\

\hline

\end{tabular}
\endgroup
}
\end{table*}

Tables~\ref{tab:attacker_ukdale} and~\ref{tab:attacker_redd} report the detailed Raw, Masked, Delta, and p-value results used to generate Fig.~\ref{fig:main_results}.

\FloatBarrier

\end{document}